\documentclass{style}

\usepackage{graphicx}
\usepackage{wrapfig}
\usepackage{booktabs}
\usepackage{multirow}
\usepackage{subcaption}
\usepackage{makecell}
\usepackage{url}
\usepackage{siunitx}
\usepackage{ulem}
\usepackage{enumitem}
\usepackage{quoting}
\usepackage{amsfonts}
\usepackage{float}
\usepackage{stfloats}
\usepackage{pifont}

\definecolor{highlightcolor}{gray}{0.95}
\tcbuselibrary{breakable}
\usepackage[svgnames]{xcolor}
\usepackage{textcase} % 提供 \MakeTextUppercase
\tcbset{
  prompt/.style={
    colback=OldLace,
    colframe=YellowGreen,
    width=\linewidth,
    arc=2mm,
    auto outer arc,
    breakable,
    lefttitle=0pt,      % 标题从最左开始（不缩进）
    boxsep=1.5mm,       % 内容与边框间距（按需调）
    left=1.5mm, right=1.5mm, top=1.5mm, bottom=1.5mm,
    fonttitle=\sffamily,  % 标题字体
    fontupper=\ttfamily,     % 内容字体
    title={\MakeTextUppercase{#1}},
  }
}

\usepackage[utf8]{inputenc}
\usepackage{longtable} % 支持长表格
\usepackage{multirow}  % 支持合并行
\usepackage{array}     % 用于自定义列格式
\usepackage[table]{xcolor} % 提供 \rowcolor
\usepackage{pifont}% http://ctan.org/pkg/pifont
\setlist[itemize]{leftmargin=*}
\definecolor{lightgray}{gray}{0.92}
\newcolumntype{C}[1]{>{\centering\arraybackslash}m{#1}}

\usepackage{times}
\usepackage{wrapfig}
\usepackage{booktabs}
\usepackage{array}
\usepackage{colortbl}
\usepackage{latexsym}
\usepackage{indentfirst}
\usepackage[T1]{fontenc}
\usepackage[utf8]{inputenc}

\usepackage{microtype}

\usepackage{inconsolata}
\usepackage{marvosym} % for \Letter (envelope symbol for corresponding author)
\usepackage{times}
\usepackage{latexsym}

\usepackage{makecell}
\usepackage[most]{tcolorbox}
\usepackage{hyperref}
\usepackage{url}
\usepackage{multirow}
\usepackage{booktabs}
\usepackage{siunitx}
\usepackage{ulem}
\usepackage{enumitem}
\usepackage{caption}
\usepackage{graphicx}
\usepackage{subcaption}
\usepackage{xcolor}
\usepackage{quoting}
\usepackage{amsmath} % For math notation if needed
\usepackage{amsfonts}
\usepackage{xcolor} % If using colored text
\usepackage{caption} % For caption spacing control
\usepackage{float} % For figure placement options
\usepackage{stfloats}
\usepackage{hyperref}
\usepackage{wrapfig}

\usepackage{multirow}
\usepackage{pifont}
\usepackage[table]{xcolor}
\usepackage[T1]{fontenc}
\usepackage[utf8]{inputenc}

\usepackage[table]{xcolor} % 提供 \rowcolor
\usepackage{pifont}% http://ctan.org/pkg/pifont
\setlist[itemize]{leftmargin=*}
\usepackage[utf8]{inputenc}

\usepackage{microtype}

\usepackage{inconsolata}

\usepackage{colortbl}
\usepackage{xcolor}
\usepackage{booktabs}
\usepackage{multirow}
\usepackage{makecell}
\usepackage{array}

\definecolor{lightred}{RGB}{250, 235, 230} % 极淡的红色用于背景
\definecolor{highlightbold}{RGB}{180, 0, 0} % 深红色用于字体
\definecolor{lightyellow}{RGB}{255, 250, 194}
\definecolor{lightgray}{RGB}{235, 235, 235}

\definecolor{humancolor}{RGB}{34,139,34}
\definecolor{modelcolor}{RGB}{0,90,180}
\definecolor{casebg}{RGB}{248,248,248}

\newcolumntype{Y}{>{\centering\arraybackslash}p{1.7em}}

\newtcolorbox{casebox}[1][]{
  colback=casebg,
  colframe=black!60,
  boxrule=0.4pt,
  arc=2pt,
  left=4pt, right=4pt, top=4pt, bottom=4pt,
  fonttitle=\bfseries\small,
  title={#1}
}

\newtcolorbox{promptbox}[1][]{
  breakable,
  colback=gray!4,
  colframe=black!55,
  boxrule=0.5pt,
  arc=2pt,
  left=5pt, right=5pt, top=4pt, bottom=4pt,
  fonttitle=\bfseries\small,
  coltitle=white,
  colbacktitle=black!60,
  title={#1}
}
\usepackage{graphicx}

\definecolor{ourcolor}{RGB}{240, 82, 156}
\definecolor{tablegray}{RGB}{223, 242, 252}
\definecolor{tablegreen}{RGB}{15, 203, 150}
\definecolor{tableyellow}{RGB}{250, 242, 233}
\definecolor{tableblue}{RGB}{240, 82, 156}
\definecolor{darkpink}{RGB}{139, 14, 98}

\usepackage[table,xcdraw]{xcolor}

\title{
    \fontsize{15.5pt}{22pt}\sffamily\bfseries\color{black}
    DecoupleMix: Decoupled Ratio Search and Convex Allocation for Scalable VLM Data Recipes
}
\newcommand{\corrmark}{\text{\Letter}}

\author{
    \fontsize{12pt}{17pt}\selectfont\sffamily\color{black}
    Jiahao Xie\thanks{Equal contribution.} \quad
    Zhongbin Guo\footnotemark[1]\thanks{Work done during an internship at ByteDance Commercial AI Team.} \quad
    Qianle Wang \quad 
    Ruiqi Lu \quad
    Dongling Xiao \\
    \vspace{0.5em}
    \quad Wanxuan Sun$^\corrmark$ \quad Cheng Yang$^\corrmark$ \\[0.5em]
    \vspace{1em}
    \normalsize\sffamily\color{black} ByteDance \\
}

\email{guozhongbin66@gmail.com}
\date{\today}

\begin{document}

\maketitle

{\renewcommand{\thefootnote}{\corrmark}%
 \footnotetext{Corresponding author.}}

\thispagestyle{firstpage}

\begin{abstract}
While data curation for Vision Language Models (VLMs) is increasingly active, public practice for constructing pretraining mixtures remains largely heuristic: practitioners stack datasets that pass quality filters, set cross-domain ratios by intuition, and lack a principled, attributable criterion for admitting new data, while frontier recipes remain undisclosed. 
We formulate data construction as a \textbf{systematic mixture-optimization} problem and turn it into a reproducible engineering discipline by decoupling the mixture into two orthogonal sub-problems: \textbf{inter-class} ratios across capabilities and \textbf{intra-class} ratios within a category. 
For inter-class allocation, we use a single-variable iterative search; for intra-class composition, we apply a multidimensional, dataset-level assessment scoring Quality and Difficulty, and formulate selection as a constrained convex optimization with a diversity objective. 
The \textbf{DecoupleMix} framework delivers two critical capabilities: guiding \textit{what data to collect next} and rendering dataset validation a controlled, attributable experiment. Experiments show our approach consistently surpasses heuristic baselines. Moreover, optimal ratios discovered on small-scale proxies transfer seamlessly to larger scales without retuning. 
Using 80B additional multimodal continue-pretraining tokens, our VLM is competitive with strong open-source models trained with substantially larger multimodal budgets.

\end{abstract}

\section{Introduction}
\label{sec:intro}

Vision Language Models (VLMs) have become a central paradigm for building general-purpose multimodal systems, combining visual perception, language understanding, and cross-modal reasoning within a unified framework. Proprietary frontier models, such as GPT-5.6~\cite{gpt56model}, Seed-2.0~\cite{seedseed20}, Claude Fable 5~\cite{fable5}, and Gemini-3.5-Flash~\cite{gemini35pro}, demonstrate remarkable capabilities across a wide range of vision--language tasks. Meanwhile, the open-source community has made substantial progress with models including LLaVA series~\cite{liu2023visual,an2025llava}, Qwen3-VL~\cite{bai2025qwen3vltechnicalreport} and Molmo2~\cite{clark2026molmo2}. Nevertheless, a considerable performance gap persists between frontier proprietary models and their open-source counterparts. 
Beyond differences in architecture and scale, growing evidence highlights the \emph{Data Mixture Recipe}---the selection, composition, and allocation of training data---as a critical determinant of model capability and data efficiency~\cite{DBLP:journals/corr/abs-2405-16640,Deng_2025_CVPR,liu2025quadmix,vosoughi2025quality}. However, proprietary recipes remain largely undisclosed, while open-source pipelines still depend heavily on quality filtering, dataset stacking, and empirically chosen sampling ratios. 
This practice leads to two critical limitations:

\begin{itemize}[leftmargin=*]
    \item \textbf{Over-reliance on Quality Filtering:} Prevailing methods focus on cleaning ``noisy'' data, but lack systematic assessment of \textit{Data Difficulty} and \textit{Diversity}. 
    \item \textbf{Unsystematic Mixture Strategies:} Ratios for \textbf{inter-class} settings (e.g., Grounding vs. Caption) are typically set by intuition. For \textbf{intra-class} settings (e.g., multiple OCR datasets), the standard practice is brute-force stacking of all filtered data, ignoring redundancy.
\end{itemize}

To address these challenges, we recast the construction of continue-pretraining data as a reproducible engineering discipline. Our central insight is that the monolithic question of ``what to mix'' can be \textbf{decoupled} into two hierarchical sub-problems: \textbf{inter-class} allocation---how to split the token budget \textit{across} capability categories, and \textbf{intra-class} composition---how to mix candidate sources \textit{within} a single category. 
This mirrors industrial data organization and sidesteps the intractability of high-dimensional search. Rather than sample-level filtering, we score each candidate dataset as a whole unit, keeping optimization tractable across hundreds of sources.

Beyond producing an optimized recipe, this decoupling yields the \textbf{attributable} evaluation  of dataset admission. By holding the inter-class ratio, total budget, and intra-class strategy fixed, adding a candidate dataset becomes a single intervention whose total effect includes the resulting within-class reallocation. This avoids the broader confounding introduced by expensive ``re-stack and retrain'' ablations.

\begin{figure*}[!t]
    \centering
    \includegraphics[width=1\linewidth]{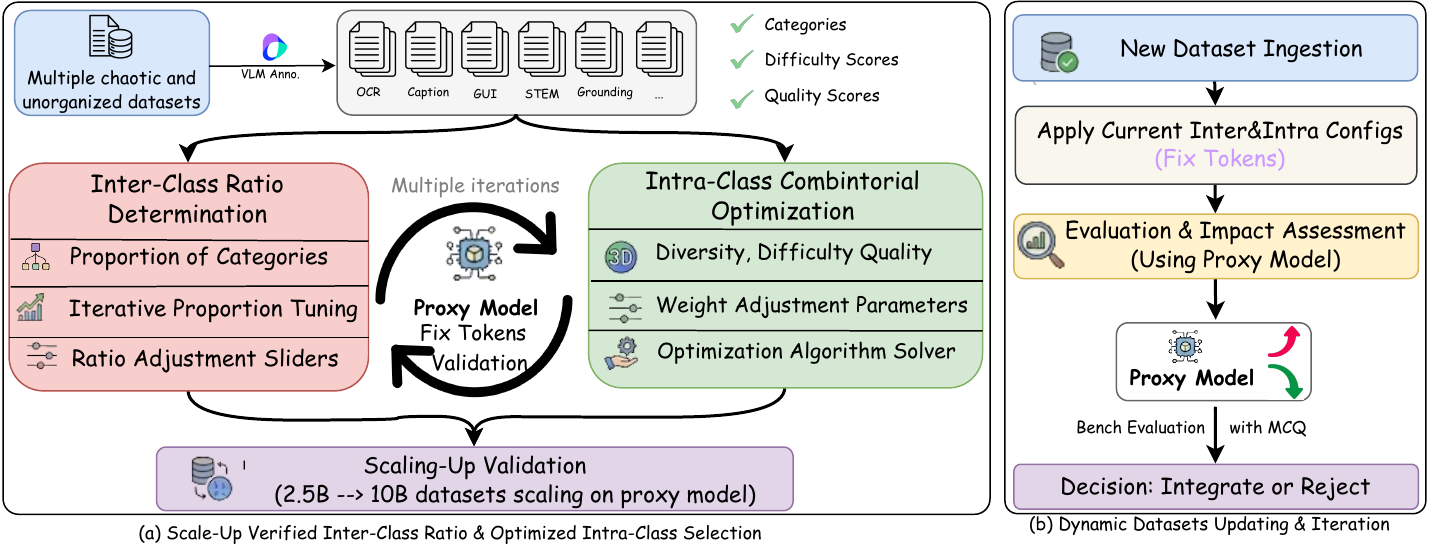}
    % \vspace{-0.8cm}
    \caption{\textbf{Overview of DecoupleMix.} \textbf{(a) Recipe search:} annotated datasets are optimized along two decoupled axes---inter-class budget allocation and intra-class convex selection---then validated and scaled up on a cheap proxy model (2.5B$\rightarrow$10B tokens). \textbf{(b) Dynamic admission:} a new dataset is evaluated under the \emph{fixed} configuration, turning its acceptance into a controlled, attributable integrate-or-reject decision.}
    \label{fig:overview}
    % \vspace{-0.8em}
\end{figure*}

Our contributions are summarized as follows:
\begin{itemize}[leftmargin=*]
    \item \textbf{Decoupled Mixture Methodology (DecoupleMix):} We decompose data construction into inter- and intra-class ratios using single-variable search and constrained convex optimization, built on an automated, multi-dimensional dataset-level assessment.
    \item \textbf{Attributable Data Acceptance:} Fixing inter-class ratios and intra-class strategies turns dataset validation into a controlled intervention whose effect can be measured with substantially less global confounding.
    \item \textbf{Scalability and Efficiency:} Ratios found at the proxy scale transfer across the data and model scales without additional tuning. With 80B additional multimodal continued-pretraining tokens, our VLM is competitive with strong open-model baselines.
\end{itemize}

\section{Methodology}
\label{sec:method}

The DecoupleMix framework forms a closed-loop system with three core modules: (1)~\textbf{Automated Dataset-Level Assessment}, which classifies capabilities and quantifies quality and difficulty at the dataset level; (2)~\textbf{Decoupled Mixture Optimization}, which uses a heuristic search for the inter-class ratio and constrained convex optimization for intra-class mixture planning; and (3)~\textbf{Standardized Data Acceptance Criteria}, which measures newly collected datasets under a fixed recipe and supports continuous iteration of the data pool. We use Multiple-Choice-Question (MCQ) validation to evaluate proxy checkpoints while reducing dependence on free-form instruction following.

\subsection{Automated Dataset-Level Assessment}

Our framework quantifies three properties: capability, quality, and difficulty. We decompose VLM capabilities into an extensible hierarchical taxonomy to identify gaps and guide data production. Its taxonomy is unbounded, and its assignment is automated: practitioners specify capability categories with natural-language criteria, and the model classifies each candidate dataset. This avoids manual labeling and allows the taxonomy to be extended when new capabilities are introduced. 

To implement this, we use datasets (rather than individual samples) as the minimal unit, which suits industrial pipelines where datasets often exhibit strong cohesion, reducing sample-level annotation noise. We adopt an LLM-as-a-Judge paradigm: we sample 150 instances per dataset and use Seed-1.6~\cite{guo2025seed1} as the evaluator. The \textbf{Quality} score $q_i$ is a weighted average of four dimensions (penalizing hallucinations), and the \textbf{Difficulty} score $d_i$ aggregates six dimensions (prioritizing cross-modal synthesis and prior knowledge). \textbf{Diversity} is not scored per dataset; instead, it is enforced structurally during mixture optimization via an entropy term (\S\ref{subsec:decoupled}). The evaluation prompts are provided in Appendix~\ref{sec:appendix_prompts}.

To confirm reliability, we also conducted a human-alignment study. Results show that automated scores are strongly correlated with human consensus (Spearman $r=0.82$ for Quality, $r=0.75$ for Difficulty), indicating that the score at dataset level is stable enough to drive downstream optimization (Appendix~\ref{subsec:appendix_human_align}).

\subsection{Decoupled Mixture Optimization}
\label{subsec:decoupled}
We solve the two decoupled dimensions with different tools: the inter-class taxonomy allocates the token budget across capability categories, while the intra-class objective captures finer-grained semantic variation \textit{within} a category through an explicit diversity term.

\paragraph{Inter-class Strategy.}
Global ratios between broad categories (e.g., Image-Text vs.\ Video-Text) shape the model's capability profile~\cite{han2026learning}. We treat these macro-distributions as hyperparameters and search them with a \textbf{coordinate-style single-variable procedure}. 
Starting from a warm-start configuration $\mathbf{r}^{(0)}$ (the average of reference recipes), each round sweeps one category's ratio $r_c$ over a candidate set while holding others fixed (renormalizing the remainder), and keeps the value maximizing validation performance. Fixing other variables isolates the effect of $c$ and avoids multi-variable confounding. We iterate until stabilization, optimizing ratios in linear rather than exponential time.

% \paragraph{Intra-class Convex Optimization.} 
% For the more granular challenge of selecting data within a category, we reject naive approaches like size-proportional sampling~\cite{liu2024llavanext} or greedy quality selection~\cite{wu2024curriculum}. We formulate intra-class sampling as a \textbf{Constrained Convex Optimization} problem, solved independently per category $c$ under its allocated budget $T_c = r_c T$.
% Within a category of $N$ datasets, we select a proportion vector $\mathbf{x} = [x_1, \ldots, x_N]^T$ to maximize a joint utility function:

% % \vspace{-0.5em}
% \begin{equation}
% \max_{\mathbf{x}} \;\sum_{i=1}^{N} \left( \alpha q_i x_i + \beta d_i x_i \right) + \gamma H(\mathbf{x})
% \end{equation}
% % \vspace{-1.2em}

% \begin{equation}
% \begin{aligned}
% \text{s.t.}\quad & L_i \le x_i \le M_i, \\
% & (1 - \tau) T_c \le \textstyle\sum_{i=1}^{N} x_i \le (1 + \tau) T_c
% \end{aligned}
% \end{equation}

% where $q_i, d_i$ are quality and difficulty scores; $H(\mathbf{x}) = -\sum_{i=1}^{N} x_i \ln x_i$ is a Shannon entropy term over the proportion vector encouraging diversity; $\alpha, \beta, \gamma$ balance the three objectives; $L_i, M_i$ are bounds; $T_c$ is the budget; and $\tau$ is tolerance.
% The entropy term prevents over-concentration on high-score datasets, forcing the ECOS solver~\cite{domahidi2013ecos} to sample from diverse, long-tail sources, ensuring a balanced, high-quality, and diverse mixture.

\paragraph{Intra-class Convex Optimization.}
For the more granular challenge of selecting data within a category, we reject naive approaches like size-proportional sampling~\cite{liu2024llavanext} or greedy quality selection~\cite{wu2024curriculum}. We formulate intra-class sampling as a \textbf{Constrained Convex Optimization} problem, solved independently per category $c$ under its allocated budget $T_c = r_c T$. Within a category of $N$ datasets, let
$\mathbf{w}=[w_1,\ldots,w_N]^\top$ denote the normalized allocation weights. We solve

\begin{equation}
\label{eq:intra_class_optimization}
\begin{aligned}
\max_{\mathbf{w}\in\Delta^{N-1}}\quad
& \sum_{i=1}^{N} w_i\bigl(\alpha q_i+\beta d_i\bigr)
  +\gamma\mathcal{H}(\mathbf{w}) \\
\mathrm{s.t.}\quad
& \ell_i\leq w_i\leq u_i,
  \qquad i=1,\ldots,N,
\end{aligned}
\end{equation}

where

\begin{equation}
\Delta^{N-1}
=
\left\{
\mathbf{w}\in\mathbb{R}_{+}^{N}
\,\middle|\,
\sum_{i=1}^{N}w_i=1
\right\},
\qquad
\mathcal{H}(\mathbf{w})
=
-\sum_{i=1}^{N}w_i\log w_i .
\end{equation}

Here, $q_i$ and $d_i$ denote the quality and difficulty scores of dataset $i$;
$\alpha,\beta,\gamma\geq0$ control their relative contributions; and
$\ell_i=L_i/T_c$ and $u_i=M_i/T_c$ are the normalized lower and upper
allocation bounds, where $L_i$ denote the corresponding token-level
bounds and
$M_i$ denotes the total number of available tokens in dataset $i$. The resulting token allocation is
\begin{equation}
t_i=T_cw_i,
\qquad
\sum_{i=1}^{N}t_i=T_c.
\end{equation}

The entropy regularizer discourages excessive concentration on a small number of high-scoring datasets and promotes broader coverage of the candidate pool. Since $\mathcal{H}(\mathbf{w})$ is concave and the feasible region is convex, Eq.~\eqref{eq:intra_class_optimization} defines a convex optimization problem, which we solve using ECOS~\cite{domahidi2013ecos}.

\subsection{Data Acceptance via Attributable Validation}
\label{subsec:attributable_method}

Validating a candidate dataset still requires an empirical training run. The fundamental challenge is to define a comparison in which admission does not simultaneously alter the global budget and inter-class composition. Our protocol creates such a controlled admission comparison, making dataset admission mathematically \textbf{identifiable}.

Let performance be $P = f(\mathbf{r}, \mathbf{x}, T)$, where $\mathbf{r}$ is the inter-class ratio, $T$ the budget, and $\mathbf{x}$ the intra-class allocation. To first order,
\begin{equation}
\mathrm{d}P = \nabla_{\mathbf{r}} f \cdot \mathrm{d}\mathbf{r} + \textstyle\sum_{c}\nabla_{\mathbf{x}_c} f \cdot \mathrm{d}\mathbf{x}_c + \tfrac{\partial f}{\partial T}\,\mathrm{d}T.
\end{equation}

In ``re-stack and retrain'', all terms move simultaneously, making $\mathrm{d}P$ confounded. Our protocol fixes $\mathbf{r}$, $T$, and the optimizer $g$, while admitting the candidate into exactly one class $c$. 
% The induced shift $\mathrm{d}\mathbf{x}_c$ is a deterministic response of the fixed optimizer to one atomic intervention. Hence the measured difference is driven by a \emph{single} controllable variable:
The induced shift $\mathrm{d}\mathbf{x}_c$ is therefore part of the response to a single admission intervention:
% \begin{equation}
% \Delta P_{d^{\star}} = f\!\big(\mathbf{r}, g(\mathcal{D}\cup\{d^{\star}\}), T\big) - f\!\big(\mathbf{r}, g(\mathcal{D}), T\big)
% \end{equation}
\begin{equation}
% \resizebox{0.85\columnwidth}{!}{$\displaystyle
\Delta P_{d^{\star}} = f\!\big(\mathbf{r}, g(\mathcal{D}\cup\{d^{\star}\}), T\big) - f\!\big(\mathbf{r}, g(\mathcal{D}), T\big)
% $}
\end{equation}

This captures the total effect of admitting the dataset---its direct contribution plus the budget-preserving re-balancing within its class. Because other factors are constant, the effect is attributable to the candidate dataset $d^{\star}$ alone. Furthermore, the convex program also provides a low-cost screening step: a source that receives zero allocation under the fixed rule need not proceed to an empirical training run.

\section{Experiments}
\label{sec:experiments}

We evaluate DecoupleMix at multiple data and model scales, separating end-to-end comparison from controlled tests of inter-class search, intra-class allocation, and dataset admission. We then scale the selected recipe to 80B additional multimodal continued-pretraining tokens for comparison with open-source model baselines.

\subsection{Experimental Setup}
\label{subsec:setup}

\paragraph{Model \& Training.} 
To obtain a controlled multimodal initialization without inheriting a pretrained cross-modal projector, we assemble our models from pretrained unimodal-stage components rather than fine-tuning an existing VLM. Following the Qwen3-VL architecture~\cite{bai2025qwen3vltechnicalreport}, we initialize the visual encoder with Qwen3-VL-ViT and the language backbone with the text-only Qwen3-4B-Instruct~\cite{yang2025qwen3} (an instruction LLM never exposed to images), and randomly initialize the ViT--LLM projector. 
All subsequent vision--language alignment and multimodal pretraining are then performed by us. Unless otherwise noted, all mixture-search, ablation, scaling, and attribution experiments are conducted on this 4B model, and the final full-scale run reuses the recipe discovered at the 4B proxy scale. Detailed hyperparameters are provided in Appendix~\ref{sec:appendix_training_details}.

\begin{table*}[t]
    \renewcommand\arraystretch{1.2} % 增加行高提升可读性
    \setlength{\tabcolsep}{3pt}     % 紧凑列间距，避免超出边距
    \centering
    \footnotesize
    \resizebox{1.0\textwidth}{!}{
    \begin{tabular}{l|c|YYYYY|YYY|YYY|YYY|YY|c}
        \toprule
        \multirow{2}{*}{\textbf{Models}} & \multirow{2}{*}{\textbf{Params.}} & \multicolumn{5}{c|}{\textbf{General}} & \multicolumn{3}{c|}{\textbf{K\&H}} & \multicolumn{3}{c|}{\textbf{OCR}} & \multicolumn{3}{c|}{\textbf{M\&L}} & \multicolumn{2}{c|}{\textbf{Video}} & \multirow{2}{*}{\textbf{Avg}} \\
        & & 
        \rotatebox{90}{MMBench}\footnotemark[1] & \rotatebox{90}{MME}\footnotemark[2] & \rotatebox{90}{MMStar} & \rotatebox{90}{R.WorldQA} & \rotatebox{90}{MMMU} & 
        \rotatebox{90}{BLINK} & \rotatebox{90}{HalBench} & \rotatebox{90}{AI2D} & 
        \rotatebox{90}{OCRBen.} & \rotatebox{90}{InfoVQA} & \rotatebox{90}{CharXiv} & 
        \rotatebox{90}{LogicVis} & \rotatebox{90}{VisuLog} & \rotatebox{90}{PuzzleVQA} & 
        \rotatebox{90}{V.MME} & \rotatebox{90}{MVBench} & 
        \\
        \midrule
        
        % ~ 2B Scale Group
        Qwen3-VL-2B-Instruct & 2B & \textbf{77.4} & \textbf{72.5} & \textbf{56.1} & \textbf{64.1} & \textbf{49.6} & \textbf{55.3} & \textbf{45.1} & \textbf{75.8} & \textbf{82.2} & \textbf{58.2} & \textbf{51.6} & \textbf{35.6} & 24.4 & 13.4 & \textbf{47.8} & 50.2 & \textbf{53.7} \\
        \textbf{Our-1B} & 1B & \cellcolor{lightred}72.2 & \cellcolor{lightred}72.0 & \cellcolor{lightred}54.2 & \cellcolor{lightred}62.1 & \cellcolor{lightred}41.0 & \cellcolor{lightred}52.9 & \cellcolor{lightred}44.6 & \cellcolor{lightred}71.2 & \cellcolor{lightred}80.4 & \cellcolor{lightred}52.9 & \cellcolor{lightred}48.6 & \cellcolor{lightred}31.5 & \cellcolor{lightred}\textbf{24.8} & \cellcolor{lightred}\textbf{30.2} & \cellcolor{lightred}45.1 & \cellcolor{lightred}\textbf{53.6} & \cellcolor{lightred}52.3 \\
        
        \arrayrulecolor{lightgray}\hline\arrayrulecolor{black}
        
        % ~ 4B Scale Group
        Qwen3-VL-4B-Instruct & 4B & \textbf{83.7} & \textbf{82.0} & \textbf{66.9} & 70.1 & \textbf{60.0} & \textbf{64.4} & 54.1 & \textbf{82.4} & \textbf{84.8} & 67.2 & 59.9 & 39.4 & \textbf{26.0} & 43.0 & \textbf{55.4} & 60.7 & 62.5 \\
        \textbf{Our-4B} & 4B & \cellcolor{lightred}83.1 & \cellcolor{lightred}80.5 & \cellcolor{lightred}66.1 & \cellcolor{lightred}\textbf{70.5} & \cellcolor{lightred}59.1 & \cellcolor{lightred}61.0 & \cellcolor{lightred}\textbf{55.8} & \cellcolor{lightred}80.7 & \cellcolor{lightred}82.7 & \cellcolor{lightred}\textbf{67.3} & \cellcolor{lightred}\textbf{63.7} & \cellcolor{lightred}\textbf{40.3} & \cellcolor{lightred}25.8 & \cellcolor{lightred}\textbf{49.0} & \cellcolor{lightred}52.9 & \cellcolor{lightred}\textbf{61.1} & \cellcolor{lightred}62.5 \\
        
        \bottomrule
    \end{tabular}
    }
    \vspace{0.8em}
    % \caption{\textbf{Comprehensive Performance Comparison.} Our models (continue-pretrained-only, no instruction tuning) vs.\ instruction-tuned Qwen3-VL-Instruct, despite being smaller (1B vs.\ 2B, 4B vs.\ 4B) and trained on far less data. Best per-benchmark in \textbf{bold}; ours in \colorbox{lightred}{light red}. Avg is the mean over all 16 benchmarks.
    \caption{\textbf{Comprehensive Performance Comparison.} Our models after multimodal pretraining and before any additional multimodal instruction tuning, compared with Qwen3-VL-Instruct. Our 1B model is smaller than the 2B baseline, while the 4B models are parameter-matched. Best score within each size group and benchmark is in \textbf{bold}; ours are shown in \colorbox{lightred}{light red}. Avg is the mean over all 16 benchmarks.
    }
    \label{tab:final_evaluation_main}
    % \vspace{-1.2em}
\end{table*}

\footnotetext[1]{MMBench/CharXiv average their sub-scores.}
\footnotetext[2]{Normalization: MME scaled to 0--100 (SUM~/~2800).}

\noindent \textbf{Training Stages:}
1) \textit{Alignment (8B tokens):} High-quality image-caption pairs align visual features with the LLM. Only the projector is trainable.
2) \textit{Mixture Validation:} Full-parameter fine-tuning on mixed datasets to assess comprehensive capabilities.
3) \textit{Annealing:} A short, high-quality refinement stage with learning rate decaying to zero.

\subsection{Evaluation Metrics}
\label{subsec:metrics}

To holistically assess model capabilities within space constraints, we use 16 benchmarks categorized into five domains matching our capability taxonomy.
These domains are: (1) General, (2) Knowledge \& Hallucination, (3) OCR \& Document, (4) Math \& Logic, (5) Video. Detailed benchmark descriptions, evaluation setups and citations are in Appendix~\ref{sec:appendix_metrics}.
For fair comparison, all reported numbers including those of external baselines are obtained by us under a single unified evaluation protocol rather than copied from their original papers; in particular, the video benchmarks are evaluated with a fixed sampling of 8 frames per video for every model.

To validate metrics without instruction-tuning alignment, for benchmarks that can be cast as multiple-choice questions we adopt Closed-Task Validation with MCQs~\cite{zhang2025closedtask}, which probes internal knowledge and reflects pretraining data efficacy by decoupling foundational understanding from instruction-following.
% benchmark descriptions, evaluation setups and citations arewith MCQs~\cite{zhang2025closedtask}, which probes internal knowledge ands
% \newpage

\section{Experiment Results and Analysis}
\label{sec:analysis}

\subsection{Main Results}
\label{subsec:main_results}

To evaluate end-to-end efficacy, we train VLMs with our pipeline by assembling a pretrained vision encoder and language model, randomly initializing their projector, and performing the subsequent multimodal alignment ourselves. We compare the resulting models with Qwen3-VL in Table~\ref{tab:final_evaluation_main}. This experiment tests whether the pipeline's end product is competitive with models trained using substantially larger multimodal budgets; controlled comparisons with heuristic stacking are reported separately in Section~\ref{subsec:scaling}.

For comparability, all numbers are obtained under the same evaluation protocol (\S\ref{subsec:metrics}). Our 1B model is compared with a larger 2B baseline, whereas the 4B comparison is parameter-matched. Both of our models use 80B additional multimodal continued-pretraining tokens and are evaluated before any additional multimodal instruction tuning (\S\ref{subsec:setup}). Under this setting, the 4B model matches Qwen3-VL-4B in overall average (\textbf{62.5} vs.\ 62.5), and the 1B model approaches Qwen3-VL-2B (\textbf{52.3} vs.\ 53.7). These results demonstrate the end-to-end competitiveness of the resulting recipe, without isolating data mixture from the other differences between systems.

\begin{wrapfigure}{r}{0.55\linewidth}
    \centering

    \renewcommand{\arraystretch}{1.15}
    \setlength{\tabcolsep}{2.5pt}
    \footnotesize

    \resizebox{\linewidth}{!}{%
    \begin{tabular}{ll|ccccc|cc}
        \toprule
        \multirow{2}{*}{\textbf{Method}} &
        \multirow{2}{*}{\textbf{Admit}} &
        \multicolumn{5}{c|}{\textbf{Per-domain $\Delta$}} &
        \multirow{2}{*}{\textbf{$\Delta$Avg}} &
        \textbf{Max} \\
        & & \textbf{OCR} & \textbf{Gen.} & \textbf{K\&H}
        & \textbf{M\&L} & \textbf{Video} & &
        $|\Delta|_{\mathrm{off}}$ \\
        \midrule
        \rowcolor{lightred}
        \textbf{Ours} & OCR set
        & +0.3 & -0.1 & +0.0 & +0.6 & +0.2
        & \textbf{+0.2} & \textbf{0.6} \\
        Stack & OCR set
        & +1.5 & -1.1 & +0.8 & -0.4 & +0.7
        & +0.1 & 1.1 \\
        Displace & OCR set
        & +0.5 & -2.3 & +0.2 & +0.9 & +0.6
        & -0.3 & 2.3 \\
        \bottomrule
    \end{tabular}%
    }

    \captionsetup{font=footnotesize,skip=4pt}
    \captionof{table}{%
        \textbf{Attributable validation of single-dataset admission.}
        Per-domain performance changes after admitting one OCR dataset.
        Our protocol produces substantially less off-target variation than
        \emph{Stack} and \emph{Displace}. Full per-benchmark results are in Table~\ref{tab:attributable_full}.
        %     \textbf{Attributable validation of single-dataset admission.} Per-domain performance change ($\Delta$) after admitting one OCR dataset under our budget-preserving, single-class protocol vs.\ two naive regimes: \emph{Stack} (concatenate and retrain, perturbing inter- and intra-class composition) and \emph{Displace} (fix the inter-class ratio but crowd out other OCR sources). Our protocol localizes the change to the target OCR domain (small off-target $|\Delta|_{\text{off}}=0.6$), whereas both naive regimes leak the perturbation into unrelated domains ($1.1$ and $2.3$).}
    }
    \label{tab:attributable}
\end{wrapfigure}

\paragraph{Where the gains concentrate.}
The per-benchmark pattern aligns with our design objective: our 4B model is strongest on several reasoning- and difficulty-sensitive benchmarks (PuzzleVQA 49.0 vs.\ 43.0; CharXiv 63.7 vs.\ 59.9; as well as RealWorldQA, HallusionBench, LogicVista, MVBench), while trailing slightly on several perception-heavy benchmarks (OCRBench, MME and AI2D).
A similar asymmetry appears at 1B, where our model surpasses the larger Qwen3-VL-2B on PuzzleVQA ($+16.8$) and MVBench ($+3.4$). 
Because the external baselines differ in data, scale, and training, this comparison is descriptive rather than causal. The matched-budget scaling study and intra-class ablation (\S\ref{subsec:scaling}, \S\ref{subsec:ablation}) provide the controlled support for the recipe design.

% A granular inspection reveals that Our 4B model significantly outpaces its direct counterpart in demanding domains that require deep reasoning and cross-modal synthesis, establishing a commanding lead in \textbf{Math \& Logic} tasks (e.g., MathVista: 71.5 vs. 63.3; DynaMath: 44.6 vs. 41.4) and complex real-world scenarios (R.WorldQA: 71.9 vs. 67.3). 
% These results empirically prove that prioritizing data diversity and difficulty over sheer volume allows industrial-grade VLMs to reliably punch above their weight class.

\subsection{Attributable Validation: Single-Dataset Admission}
\label{subsec:attributable}
% A central claim is that admitting a dataset under our protocol yields an effect \textit{attributable} to that dataset, avoiding confounded global shifts. We test this directly by admitting one OCR dataset under three regimes and measuring per-domain performance change $\Delta$: \textbf{Ours} (inter-class ratio $\mathbf{r}$ and total budget $T$ fixed, candidate enters via convex intra-class allocation); \textbf{Stack} (candidate concatenated, perturbing inter- and intra-class composition); \textbf{Displace} ($\mathbf{r}$ fixed, but candidate naively crowds out other OCR sources).

A central claim is that admitting a dataset under our protocol yields an effect \textit{attributable} to that dataset, avoiding confounded global shifts. We test whether the admission protocol reduces global confounding by admitting one OCR dataset under three regimes and measuring per-domain performance change $\Delta$: \textbf{Ours} (inter-class ratio $\mathbf{r}$ and total budget $T$ fixed, with the candidate entering through convex intra-class reallocation); \textbf{Stack} (the candidate is concatenated, changing the effective budget and mixture); and \textbf{Displace} ($\mathbf{r}$ is fixed, but the candidate naively crowds out other OCR sources).

\begin{table*}[t]
    \centering
    \renewcommand\arraystretch{1.2}
    \setlength{\tabcolsep}{2.5pt}
    \footnotesize
    \resizebox{\textwidth}{!}{
    \begin{tabular}{l|c|YYYYY|YYY|YYY|YYY|YY|c}
        \toprule
        \multirow{2}{*}{\textbf{Models}} & \multirow{2}{*}{\textbf{Scale}} & \multicolumn{5}{c|}{\textbf{General}} & \multicolumn{3}{c|}{\textbf{K\&H}} & \multicolumn{3}{c|}{\textbf{OCR}} & \multicolumn{3}{c|}{\textbf{M\&L}} & \multicolumn{2}{c|}{\textbf{Video}} & \multirow{2}{*}{\textbf{Avg}} \\
        & & 
        \rotatebox{90}{MMBench} & \rotatebox{90}{MME} & \rotatebox{90}{MMStar} & \rotatebox{90}{R.WorldQA} & \rotatebox{90}{MMMU} & 
        \rotatebox{90}{BLINK} & \rotatebox{90}{HalBench} & \rotatebox{90}{AI2D} & 
        \rotatebox{90}{OCRBench} & \rotatebox{90}{InfoVQA} & \rotatebox{90}{CharXiv} & 
        \rotatebox{90}{LogicVista} & \rotatebox{90}{VisuLogic} & \rotatebox{90}{PuzzleVQA} & 
        \rotatebox{90}{V.MME} & \rotatebox{90}{MVBench} & 
        \\
        \midrule
        Baseline & \multirow{2}{*}{2.5B} & 79.5 & 72.8 & 58.4 & \textbf{66.7} & 51.0 & 54.6 & 66.0 & 76.0 & 78.2 & 58.5 & \textbf{53.1} & 32.9 & 25.1 & 29.9 & 51.5 & \textbf{52.0} & 56.6 \\
        \textbf{Ours} & & \cellcolor{lightred}\textbf{82.3} & \cellcolor{lightred}\textbf{73.2} & \cellcolor{lightred}\textbf{60.7} & \cellcolor{lightred}66.5 & \cellcolor{lightred}\textbf{52.2} & \cellcolor{lightred}\textbf{55.5} & \cellcolor{lightred}\textbf{68.0} & \cellcolor{lightred}\textbf{78.0} & \cellcolor{lightred}\textbf{79.4} & \cellcolor{lightred}\textbf{60.6} & \cellcolor{lightred}52.6 & \cellcolor{lightred}\textbf{36.2} & \cellcolor{lightred}\textbf{23.1} & \cellcolor{lightred}\textbf{38.9} & \cellcolor{lightred}\textbf{51.8} & \cellcolor{lightred}51.7 & \cellcolor{lightred}\textbf{58.2} \\
        \arrayrulecolor{lightgray}\hline\arrayrulecolor{black}
        
        Baseline & \multirow{2}{*}{5B} & 80.7 & 72.6 & 59.1 & 66.3 & 51.0 & 55.5 & 68.8 & \textbf{79.7} & 77.6 & 61.7 & 53.4 & 33.6 & 22.6 & \textbf{43.8} & \textbf{52.8} & 51.9 & 58.2 \\
        \textbf{Ours} & & \cellcolor{lightred}\textbf{82.5} & \cellcolor{lightred}\textbf{74.6} & \cellcolor{lightred}\textbf{62.5} & \cellcolor{lightred}\textbf{69.3} & \cellcolor{lightred}\textbf{51.9} & \cellcolor{lightred}\textbf{57.0} & \cellcolor{lightred}\textbf{69.3} & \cellcolor{lightred}81.2 & \cellcolor{lightred}\textbf{81.9} & \cellcolor{lightred}\textbf{64.8} & \cellcolor{lightred}\textbf{55.5} & \cellcolor{lightred}\textbf{38.0} & \cellcolor{lightred}\textbf{26.4} & \cellcolor{lightred}42.4 & \cellcolor{lightred}52.5 & \cellcolor{lightred}\textbf{52.6} & \cellcolor{lightred}\textbf{60.2} \\
        \arrayrulecolor{lightgray}\hline\arrayrulecolor{black}
        
        Baseline & \multirow{2}{*}{10B} & 80.9 & \textbf{78.8} & 58.8 & 68.0 & 53.8 & 55.3 & 68.0 & 78.2 & 80.1 & 64.5 & 57.0 & 35.3 & \textbf{24.6} & 45.6 & 52.4 & \textbf{54.0} & 59.7 \\
        \textbf{Ours} & & \cellcolor{lightred}\textbf{83.2} & \cellcolor{lightred}77.6 & \cellcolor{lightred}\textbf{63.1} & \cellcolor{lightred}\textbf{69.0} & \cellcolor{lightred}\textbf{54.7} & \cellcolor{lightred}\textbf{58.5} & \cellcolor{lightred}\textbf{68.9} & \cellcolor{lightred}\textbf{80.1} & \cellcolor{lightred}\textbf{81.6} & \cellcolor{lightred}\textbf{67.6} & \cellcolor{lightred}\textbf{57.3} & \cellcolor{lightred}\textbf{36.0} & \cellcolor{lightred}{24.3} & \cellcolor{lightred}\textbf{48.6} & \cellcolor{lightred}\textbf{53.7} & \cellcolor{lightred}{52.9} & \cellcolor{lightred}\textbf{61.1} \\
        \bottomrule
    \end{tabular}
    }
    \caption{\textbf{Comprehensive Scaling Analysis.} We compare our full method (\textbf{Ours}: optimized inter-class ratios with convex intra-class allocation) against the \textbf{Baseline} (heuristic stacking: all quality-filtered datasets pooled and sampled in proportion to their size) across three \emph{training-token} scales (2.5B, 5B, 10B). 
    % We report the mean score over each of the five capability domains plus the overall average. 
    % Full results are in Table~\ref{tab:scaling_full}.
    }
    \label{tab:scaling_main}
\end{table*}

\begin{table*}[t]
    \centering
    \renewcommand\arraystretch{1.2}
    \setlength{\tabcolsep}{2.5pt}
    \footnotesize
    \resizebox{\textwidth}{!}{
    \begin{tabular}{l|YYYYY|YYY|YYY|YYY|YY|c}
        \toprule
        \multirow{2}{*}{\textbf{32B Recipe}} & \multicolumn{5}{c|}{\textbf{General}} & \multicolumn{3}{c|}{\textbf{K\&H}} & \multicolumn{3}{c|}{\textbf{OCR}} & \multicolumn{3}{c|}{\textbf{M\&L}} & \multicolumn{2}{c|}{\textbf{Video}} & \multirow{2}{*}{\textbf{Avg}} \\
        & 
        \rotatebox{90}{MMBench} & \rotatebox{90}{MME} & \rotatebox{90}{MMStar} & \rotatebox{90}{R.WorldQA} & \rotatebox{90}{MMMU} & 
        \rotatebox{90}{BLINK} & \rotatebox{90}{HalBench} & \rotatebox{90}{AI2D} & 
        \rotatebox{90}{OCRBench} & \rotatebox{90}{InfoVQA} & \rotatebox{90}{CharXiv} & 
        \rotatebox{90}{LogicVista} & \rotatebox{90}{VisuLogic} & \rotatebox{90}{PuzzleVQA} & 
        \rotatebox{90}{V.MME} & \rotatebox{90}{MVBench} & 
        \\
        \midrule
        Stacking & 83.8 & 80.8 & 63.9 & 69.3 & 57.3 & 55.9 & 50.0 & \textbf{83.4} & 79.0 & 62.4 & 59.2 & 40.9 & 24.0 & \textbf{60.4} & \textbf{57.2} & 56.3 & 61.5 \\
        \textbf{Ours} & \cellcolor{lightred}\textbf{85.3} & \cellcolor{lightred}\textbf{82.2} & \cellcolor{lightred}\textbf{66.1} & \cellcolor{lightred}\textbf{70.1} & \cellcolor{lightred}\textbf{59.4} & \cellcolor{lightred}\textbf{59.3} & \cellcolor{lightred}\textbf{51.9} & \cellcolor{lightred}81.8 & \cellcolor{lightred}\textbf{82.9} & \cellcolor{lightred}\textbf{63.5} & \cellcolor{lightred}\textbf{60.6} & \cellcolor{lightred}\textbf{42.5} & \cellcolor{lightred}\textbf{25.9} & \cellcolor{lightred}56.8 & \cellcolor{lightred}57.1 & \cellcolor{lightred}\textbf{57.4} & \cellcolor{lightred}\textbf{62.7} \\
        \bottomrule
    \end{tabular}
    }
    % \caption{\label{tab:scale_32b_full}
    % \textbf{Full per-benchmark 32B transfer results.} Our proxy-searched recipe (no re-tuning) vs.\ heuristic stacking at a matched token budget on a 32B model. Better score per column in \textbf{bold}.
    % }
    \caption{\label{tab:scale_32b}
    \textbf{Cross-model-scale transfer to 32B.} Full per-benchmark comparison of our proxy-searched recipe, applied without retuning, with heuristic stacking at a matched token budget. Best score per column is in \textbf{bold}; ours are shown in \colorbox{lightred}{light red}.
    % ; the full per-benchmark breakdown is in Table~\ref{tab:scale_32b_full}.
    }
\end{table*}

\paragraph{Result.}
Table~\ref{tab:attributable} reports the outcome. Under our protocol, the response is localized: the target OCR domain moves by $+0.3$ while every off-target domain stays within $0.6$. Naive regimes leak perturbation: stacking drifts the global balance (max off-target $1.1$), and displacement regresses General by $-2.3$. 
Stacking shows a larger raw OCR increase ($+1.5$) because it assigns a larger share of the resulting global mixture to OCR, conflating source admission with global rebalancing. Despite this change, its overall average increases by only $+0.1$, compared with $+0.2$ under our budget-preserving protocol; displacement decreases the average by $-0.3$.
The directions of change are also coherent under our protocol: Math \& Logic, which can depend on OCR-style perception, improves by $+0.6$, whereas stacking changes it by $-0.4$ as part of the broader mixture shift. This comparison turns ``should this source be ingested?'' into a controlled policy-level decision (full per-benchmark scores in Appendix Table~\ref{tab:attributable_full}). The transfer results in \S\ref{subsec:scaling} support reusing the surrounding recipe at larger scales, although the admission effect itself is measured only at the proxy scale.

\subsection{Scaling Efficiency and Transferability}
\label{subsec:scaling}
We analyze our full method across data scales (2.5B, 5B, 10B tokens) in Table~\ref{tab:scaling_main}.

\paragraph{Consistent Superiority.} 
Our method outperforms the baseline at every scale: the average gap is $+1.6$ at 2.5B, $+2.0$ at 5B, and $+1.4$ at 10B. The advantage is stable rather than monotonically growing, maintaining a consistent edge. Improvements are pervasive across capability domains (e.g., InfoVQA $+3.1$ at 10B).

\begin{table*}[t]
    \centering
    \renewcommand\arraystretch{1.2}
    \setlength{\tabcolsep}{2.5pt}
    \footnotesize
    \resizebox{\textwidth}{!}{
    \begin{tabular}{l|YYYYY|YYY|YYY|YYY|YY|c}
        \toprule
        \multirow{2}{*}{\textbf{Ablation Settings}} & \multicolumn{5}{c|}{\textbf{General}} & \multicolumn{3}{c|}{\textbf{K\&H}} & \multicolumn{3}{c|}{\textbf{OCR}} & \multicolumn{3}{c|}{\textbf{M\&L}} & \multicolumn{2}{c|}{\textbf{Video}} & \multirow{2}{*}{\textbf{Avg}} \\
        & 
        \rotatebox{90}{MMBench} & \rotatebox{90}{MME} & \rotatebox{90}{MMStar} & \rotatebox{90}{R.WorldQA} & \rotatebox{90}{MMMU} & 
        \rotatebox{90}{BLINK} & \rotatebox{90}{HalBench} & \rotatebox{90}{AI2D} & 
        \rotatebox{90}{OCRBench} & \rotatebox{90}{InfoVQA} & \rotatebox{90}{CharXiv} & 
        \rotatebox{90}{LogicVista} & \rotatebox{90}{VisuLogic} & \rotatebox{90}{PuzzleVQA} & 
        \rotatebox{90}{V.MME} & \rotatebox{90}{MVBench} & 
        \\
        \midrule
        \textbf{Baseline (Stacking)} & 79.5 & 72.8 & 58.4 & \textbf{66.7} & 51.0 & 54.6 & 66.0 & 76.0 & 78.2 & 58.5 & \textbf{53.1} & 32.9 & 25.1 & 29.9 & 51.5 & 52.0 & 56.6 \\
        \midrule
        \multicolumn{18}{l}{\textit{Inter-class Ratio Variants (Intra-class fixed as Baseline)}} \\
        $\bullet$ Bee Recipe & 77.9 & 71.8 & 56.4 & 64.4 & \textbf{56.2} & 54.1 & 67.5 & 78.1 & 78.4 & 60.6 & 50.5 & 34.2 & \textbf{25.6} & 20.4 & 50.5 & 51.2 & 56.1 \\
        $\bullet$ LLaVA-OV 1.5 Recipe & 79.5 & \textbf{73.6} & 58.9 & \textbf{66.7} & 52.7 & 55.8 & 65.1 & \textbf{79.0} & 77.5 & 59.1 & 52.0 & 34.5 & 24.1 & 30.9 & 51.7 & 50.8 & 57.0 \\
        $\bullet$ \textbf{Ours (Inter-only)} & 79.7 & 72.3 & 57.7 & 66.0 & 50.4 & \textbf{56.5} & 66.8 & 77.1 & 77.0 & 58.5 & 51.5 & 30.6 & 27.2 & 37.5 & 51.7 & \textbf{52.7} & 57.1 \\
        \midrule
        \rowcolor{lightred}
        \textbf{Ours (Inter + Intra)} & \textbf{82.3} & 73.2 & \textbf{60.7} & 66.5 & 52.2 & 55.5 & \textbf{68.0} & 78.0 & \textbf{79.4} & \textbf{60.6} & \textbf{52.6} & \textbf{36.2} & 23.1 & \textbf{38.9} & \textbf{51.7} & 51.7 & \textbf{58.2} \\
        \bottomrule
    \end{tabular}
    }
    % \caption{\label{tab:ablation_full}
    % \textbf{Full per-benchmark decoupling ablation.} Per-benchmark breakdown for the ablation in Table~\ref{tab:ablation}. Best per column in \textbf{bold}; full method in \colorbox{lightred}{light red}.
    % }
    \caption{\label{tab:ablation}
    \textbf{Ablation on Decoupled Strategies.}
    We compare our optimization against baseline stacking and alternative inter-class recipes (Bee, LLaVA-OV-1.5), reporting all per-benchmark scores and overall average. Best per column in \textbf{bold}; full method in \colorbox{lightred}{light red}.
    % Full results in Table~\ref{tab:ablation_full}.
    }
\end{table*}

\paragraph{Transferability Across Data and Model Scales.}
Optimal ratios searched only at 2.5B transfer to 5B and 10B scales without retuning and outperform the corresponding baselines. This indicates that the proxy search captures regularities that persist across the data scales.
To probe transferability across model sizes, we evaluated the same 2.5B proxy recipe on a 32B model without retuning (Table~\ref{tab:scale_32b}). The transferred recipe improves the overall average by $+1.2$ (61.5 $\rightarrow$ 62.7) against a matched-budget stacking baseline and leads on 13 of 16 benchmarks (e.g., OCR $+2.1$). 
This reinforces that discovered ratios reflect universal data properties rather than model-specific artifacts.

\paragraph{Search cost is amortized at the proxy scale.}
This transferability is what makes the method economical in practice. The single-variable inter-class search is conducted at the 2.5B-token proxy scale, where each candidate run has lower cost; the resulting recipe is then reused unchanged across both larger data budgets (5B, 10B) and larger model capacities (32B). 
Within these experiments, the search cost is paid once and amortized across downstream runs, unlike ``re-stack and retune'' workflows that repeat the sweep for each configuration.

\subsection{Ablation: Decoupling and Intra-class Allocation}
\label{subsec:ablation}

To attribute our gains, we decompose the effect of inter-class and intra-class optimization (Table~\ref{tab:ablation}), and analyze the intra-class allocator itself (Table~\ref{tab:intra_class_ablation}).

% \begin{table}[t]
%     \centering
%     \footnotesize
%     \renewcommand\arraystretch{1.2}
%     \setlength{\tabcolsep}{4pt}
%     \resizebox{\columnwidth}{!}{
%     \begin{tabular}{l|ccccc|c}
%         \toprule
%         \textbf{32B} & \textbf{Gen.} & \textbf{K\&H} & \textbf{OCR} & \textbf{M\&L} & \textbf{Video} & \textbf{Avg} \\
%         \midrule
%         Stacking & 71.0 & 63.1 & 66.9 & \textbf{41.8} & 56.8 & 61.5 \\
%         \rowcolor{lightred}
%         \textbf{Ours} & \textbf{72.6} & \textbf{64.3} & \textbf{69.0} & 41.7 & \textbf{57.2} & \textbf{62.7} \\
%         \bottomrule
%     \end{tabular}
%     }
%     \caption{\label{tab:scale_32b}
%     \textbf{Cross-model-scale transfer to 32B.} Our proxy-searched recipe, applied without re-tuning, vs.\ heuristic stacking at a matched token budget, reported as per-domain averages over five capability domains; the full per-benchmark breakdown is in Table~\ref{tab:scale_32b_full}.
%     }
%     % \vspace{-1.5em}
% \end{table}

\begin{table*}[t]
    \centering
    \renewcommand\arraystretch{1.2}
    \setlength{\tabcolsep}{2.5pt}
    \footnotesize
    \resizebox{\textwidth}{!}{
    \begin{tabular}{l|YYYYY|YYY|YYY|YYY|YY|c}
        \toprule
        \multirow{2}{*}{\textbf{Intra-class Strategy}} & \multicolumn{5}{c|}{\textbf{General}} & \multicolumn{3}{c|}{\textbf{K\&H}} & \multicolumn{3}{c|}{\textbf{OCR}} & \multicolumn{3}{c|}{\textbf{M\&L}} & \multicolumn{2}{c|}{\textbf{Video}} & \multirow{2}{*}{\textbf{Avg}} \\
        & 
        \rotatebox{90}{MMBench} & \rotatebox{90}{MME} & \rotatebox{90}{MMStar} & \rotatebox{90}{R.WorldQA} & \rotatebox{90}{MMMU} & 
        \rotatebox{90}{BLINK} & \rotatebox{90}{HalBench} & \rotatebox{90}{AI2D} & 
        \rotatebox{90}{OCRBench} & \rotatebox{90}{InfoVQA} & \rotatebox{90}{CharXiv} & 
        \rotatebox{90}{LogicVista} & \rotatebox{90}{VisuLogic} & \rotatebox{90}{PuzzleVQA} & 
        \rotatebox{90}{V.MME} & \rotatebox{90}{MVBench} & 
        \\
        \midrule
        Quantity-Proportional & 63.8 & 72.9 & 42.7 & 53.1 & 39.1 & 44.2 & 50.3 & 60.9 & 67.0 & 31.5 & 35.8 & 30.6 & 23.7 & \textbf{32.9} & \textbf{45.6} & 42.1 & 46.0 \\
        Quality Only & 66.3 & 71.5 & 44.4 & \textbf{57.4} & \textbf{39.2} & 43.9 & 51.9 & \textbf{65.1} & 70.4 & 36.9 & 36.5 & \textbf{31.1} & 23.4 & 26.5 & 45.0 & 43.7 & 47.1 \\
        \rowcolor{lightred}
        \textbf{Convex Alloc. (Ours)} & \textbf{67.6} & \textbf{74.8} & \textbf{45.0} & 56.5 & 38.9 & \textbf{44.9} & \textbf{52.1} & 64.9 & \textbf{73.0} & \textbf{47.1} & \textbf{38.0} & 30.0 & \textbf{25.6} & 30.8 & 45.5 & \textbf{44.6} & \textbf{48.7} \\
        \bottomrule
    \end{tabular}
    }
    % \caption{\label{tab:intra_class_ablation_full}
    % \textbf{Full per-benchmark intra-class ablation.} Per-benchmark breakdown for the strategies in Table~\ref{tab:intra_class_ablation}. Best per column in \textbf{bold}; ours in \colorbox{lightred}{light red}.}
    \caption{\label{tab:intra_class_ablation}
    \textbf{Intra-class allocation under a fixed inter-class budget.} Full per-benchmark comparison of three sampling strategies under a 1B-token budget. Best score per column is in \textbf{bold}; ours are shown in \colorbox{lightred}{light red}.
    % Full breakdown in Table~\ref{tab:intra_class_ablation_full}.
}
\end{table*}

% \vspace{-0.5em}
\paragraph{Inter-class ratios provide a strong macro-structure.}
We compare our inter-class proportions against baseline stacking and two published reference recipes (Bee~\cite{zhang2025bee} and LLaVA-OneVision-1.5~\cite{an2025llava}). Our Inter-only strategy (Avg 57.1), compared with 56.6 for stacking and 56.1 and 57.0 for the reference recipes, showing our capability-driven taxonomy allocates data better than empirical structures. Adding intra-class allocation further raises the average to 58.2.

% \vspace{-0.5em}
\paragraph{Balanced intra-class allocation beats naive sampling.}
We compare our convex allocation against two baselines under a fixed inter-class budget (Table~\ref{tab:intra_class_ablation}): \textit{Quantity-Proportional} (suboptimal Avg 46.0) and \textit{Quality-Only} (improves to 47.1 overall but underperforms on Math \& Logic, 27.0, by over-concentrating on easy high-quality data). 
Our convex allocation attains the highest average (\textbf{48.7}), with its largest domain-level improvement on OCR (52.7 vs.\ 44.8). 
This pattern is consistent with the entropy regularizer reducing concentration on a small set of high-scoring datasets and preserves long-tail coverage, as visualized in Fig.~\ref{fig:qdd_scatter}.

\section{Related Works}

\paragraph{Multimodal data curation.}
Large-scale VLM pipelines commonly improve data efficiency through similarity-based filtering~\cite{radford2021learning}, model-based quality curation~\cite{dong-etal-2025-scalable}, and joint quality--diversity selection~\cite{liu2025quadmix}. These approaches primarily score or retain individual examples. DecoupleMix is complementary: it treats datasets as allocation units after curation and addresses how the retained sources should share a finite training budget.

\paragraph{Data-mixture optimization.}
Data-mixture optimization has been studied extensively in language-model pretraining. DoReMi learns domain weights with a proxy model and group distributionally robust optimization~\cite{xie2023doremi}; Data Mixing Laws and RegMix use small-scale training runs to predict the performance of unseen mixtures~\cite{ye2024datamixinglaws,liu2024regmix}; and RegMix-D extends regression-based mixture optimization to dynamic schedules~\cite{zhao2026regmixd}. Recent work also studies scaling laws for optimal mixtures across language, vision, and native multimodal settings~\cite{shukor2025scaling}. These studies establish the broader value of mixture optimization and motivate our VLM-specific hierarchical formulation.

\paragraph{Multimodal mixture recipes.}
MM1 demonstrates that the proportions of caption, interleaved image--text, and text-only data materially affect multimodal pretraining~\cite{mckinzie2024mm1}, while open pipelines such as LLaVA-OneVision-1.5 and Bee provide reproducible reference recipes~\cite{an2025llava,zhang2025bee}. DecoupleMix differs by jointly separating inter-class budget search from intra-class dataset allocation and by evaluating candidate admission under a fixed budget and allocation rule. The contribution is therefore not the first use of mixture optimization, but a dataset-level, hierarchical workflow designed for continuously evolving VLM data pools.

\section{Limitations}

Several limitations remain and point to future work. \textbf{(1) Scaling scope:} our study spans 2.5B--10B tokens with model-scale transfer evaluated at 32B parameters; we have not released the trend at the much larger token budgets of frontier industrial pretraining. \textbf{(2) Optimization algorithms:} the single-variable inter-class search and convex intra-class program are deliberately tractable but likely suboptimal---richer joint search or more expressive objectives could improve the recipes. \textbf{(3) Modality coverage:} we validate only on vision--language models, leaving the extension to fully omni-modal models (audio, video, etc.) untested. Finally, our judge-based assessment is bounded by the frozen judge's capability, and treating each dataset as an atomic quality--difficulty unit precludes finer sample-level gains.

% \begin{table}[t]
%     \centering
%     \renewcommand\arraystretch{1.2}
%     \setlength{\tabcolsep}{4pt}
%     \footnotesize
%     \resizebox{\columnwidth}{!}{
%     \begin{tabular}{l|ccccc|c}
%         \toprule
%         \textbf{Ablation Settings} & \textbf{Gen.} & \textbf{K\&H} & \textbf{OCR} & \textbf{M\&L} & \textbf{Video} & \textbf{Avg} \\
%         \midrule
%         \textbf{Baseline (Stacking)} & 65.7 & 65.5 & 63.3 & 29.3 & 51.8 & 56.6 \\
%         \midrule
%         \multicolumn{7}{l}{\textit{Inter-class Ratio Variants (Intra fixed as Baseline)}} \\
%         $\bullet$ Bee Recipe & 65.3 & 66.6 & 63.2 & 26.7 & 50.9 & 56.1 \\
%         $\bullet$ LLaVA-OV 1.5 Recipe & 66.3 & 66.6 & 62.9 & 29.8 & 51.3 & 57.0 \\
%         $\bullet$ \textbf{Ours (Inter-only)} & 65.2 & 66.8 & 62.3 & 31.8 & \textbf{52.2} & 57.1 \\
%         \midrule
%         \rowcolor{lightred}
%         \textbf{Ours (Inter + Intra)} & \textbf{67.0} & \textbf{67.2} & \textbf{64.2} & \textbf{32.7} & 51.7 & \textbf{58.2} \\
%         \bottomrule
%     \end{tabular}
%     }
%     \caption{\label{tab:ablation}
%     \textbf{Ablation on Decoupled Strategies.}
%     We compare our optimization against baseline stacking and alternative inter-class recipes (Bee, LLaVA-OV-1.5), reporting domain means and overall average. Full results in Table~\ref{tab:ablation_full}.
%     }
%     % \vspace{-1.5em}
% \end{table}

\section{Conclusion}
We present \textbf{DecoupleMix}, a systematic data mixture optimization framework that transforms VLM pretraining from trial-and-error into principled engineering. By introducing a multi-dimensional dataset-level assessment and decoupling data composition into inter-class and intra-class optimization, we establish a scalable methodology for constructing high-performance recipes. Validated across 2.5B to 10B tokens and transferring without retuning to larger data and model scales, our approach consistently outperforms heuristic baselines, demonstrating that systematic data curation is the critical lever for efficient VLM development.

\bibliographystyle{unsrtnat}
\bibliography{custom}

\newpage

\appendix

\section{Detailed Evaluation Metrics}
\label{sec:appendix_metrics}

To comprehensively evaluate the performance of our models across diverse multimodal capability dimensions, we utilize a robust suite of 16 mainstream benchmarks. These are categorized into five primary domains, directly aligning with the capability taxonomy discussed in our methodology:

\textbf{General.} 
This domain assesses broad multimodal understanding, real-world knowledge, and complex reasoning capabilities. The evaluation suite includes \textbf{MMBench}~\cite{liu2024mmbench}, \textbf{MME}~\cite{fu2025mme}, \textbf{MMStar}~\cite{chen2024we}, \textbf{MMMU}~\cite{yue2024mmmu} (focusing on expert-level multi-discipline knowledge), and \textbf{RealWorldQA}.

\textbf{Knowledge \& Hallucination.} 
To evaluate fine-grained visual perception, diagram/knowledge understanding, and the model's robustness against hallucinating non-existent visual elements or being misled by text prompts, we employ \textbf{BLINK}~\cite{fu2024blink}, \textbf{HallusionBench}~\cite{guan2024hallusionbench}, and \textbf{AI2D}~\cite{kembhavi2016diagram}.

\textbf{OCR \& Document.} 
This category focuses on text-rich image comprehension, document layout parsing, and complex chart reasoning. It incorporates \textbf{OCRBench}~\cite{liu2024ocrbench}, \textbf{InfoVQA}~\cite{mathew2022infographicvqa}, and \textbf{CharXiv}~\cite{wang2024charxiv}.

\textbf{Math \& Logic.} 
To rigorously test the model's visual logic deduction, spatial intelligence, and geometric problem-solving abilities, we utilize \textbf{LogicVista}~\cite{xiao2024logicvista}, \textbf{VisuLogic}~\cite{xu2025visulogic} and \textbf{PuzzleVQA}~\cite{chia2024puzzlevqa}.

\textbf{Video.} 
For evaluating dynamic scene understanding, temporal localization, and cross-frame synergies, we extend our evaluation to the temporal domain using \textbf{Video-MME}~\cite{fu2025video} and \textbf{MVBench}~\cite{li2024mvbench}. For both benchmarks we uniformly sample a fixed 8 frames per video and apply the same protocol to all models, including the external baselines.

\section{Dataset-Level Assessment: Prompts and Formulations}
\label{sec:appendix_prompts}

In this section, we detail the formulations and the specific system prompts used by the SOTA VLM to evaluate the Quality ($Q$) and Difficulty ($D$) of the sampled pretraining data. For each dataset, 150 samples are evaluated to estimate the overall distribution.

\subsection{Human Alignment of the Automated Judge}
\label{subsec:appendix_human_align}
To verify that the LLM-as-a-Judge scores are reliable rather than arbitrary, we conduct a human-alignment study. We randomly sample 10 instances spanning all capability categories and ask three expert annotators to independently rate each instance on the same Quality and Difficulty rubrics used by the automated judge (1--5 scale per dimension). We take the median of the three annotators as the human consensus and measure rank agreement between the automated scores and this consensus using the Spearman correlation coefficient.

\begin{table}[h]
    \centering
    \small
    \begin{tabular}{lcc}
        \toprule
        \textbf{Dimension} & \textbf{Spearman $r$} & \textbf{$P$-value} \\
        \midrule
        Quality ($Q$)    & 0.82 & 0.004 \\
        Difficulty ($D$) & 0.75 & 0.013 \\
        \bottomrule
    \end{tabular}
    \caption{\label{tab:human_align} Agreement between the automated judge and the human consensus over 10 instances scored by three expert annotators. Both correlations are statistically significant ($P<0.05$).}
\end{table}

As shown in Table~\ref{tab:human_align}, the automated scores correlate strongly and significantly with the human consensus on both dimensions ($r=0.82$ for Quality, $r=0.75$ for Difficulty), indicating that the dataset-level Quality and Difficulty scoring is stable enough to drive the downstream mixture optimization. We will release additional annotated demonstrations in the camera-ready version.

\subsection{Quality Assessment Formulation}
The final Quality score $Q$ for a single data instance is calculated using a weighted average of four dimensions. Accuracy and Hallucination are given higher weights to ensure the factual correctness of the pretraining signals. The formula is defined as:
\begin{equation}
    Q = \frac{2 \cdot S_{acc} + 1.5 \cdot S_{hal} + 1 \cdot S_{cor} + 0.3 \cdot S_{gra}}{4.8}
\end{equation}
where $S_{acc}$, $S_{hal}$, $S_{cor}$, and $S_{gra}$ represent the scores (1-5 scale) for Accuracy, Hallucination, Correlation, and Grammar, respectively.

\vspace{0.5em}
\begin{promptbox}[Quality Evaluation Prompt]
{\small \ttfamily
\noindent You are a meticulous data quality assessor for VLM training data. Analyze each data sample, which may include both image and text (user question and assistant response) modalities, across multiple quality dimensions and provide numerical scores with clear, evidence-based reasoning. Pay special attention to hallucinations—assertions in the assistant's response that are not directly supported or verifiable from the image or text context.

\noindent \textbf{1. Evaluation Dimensions} \\
Score each dimension on a 1-5 scale (1=lowest, 5=highest):
\begin{itemize}
    \vspace{-0.5em}\item \textbf{Accuracy:} Factual correctness and verifiability of the assistant's response...
    \vspace{-0.5em}\item \textbf{Grammar:} Linguistic correctness \& fluency of text.
    \vspace{-0.5em}\item \textbf{Correlation:} Relevance of the question and answer to the image content...
    \vspace{-0.5em}\item \textbf{Hallucination:} Degree of hallucinated content, i.e., claims or inferences not grounded in visible image evidence... Lower score means more hallucination.
\end{itemize}

\noindent \textbf{2. Scoring Protocol} \\
- Base scores strictly on concrete, verifiable evidence from both text and image.\\
- Score based on the percentage of errors: 5 points for completely correct, 4 points for errors within 10\%, 3 points for errors within 20\%, 2 points for errors within 35\%, and 1 point for errors at 40\% or above.\\
- Flag hallucinations, misinformation, or unsupported inferences explicitly.\\
- Mark samples needing human review if confidence <90\%.

\noindent \textbf{3. Output Format} \\
Respond with a JSON dictionary:
\{
  "dimension\_scores": \{
    "accuracy": <int>,
    "grammar": <int>,
    "correlation": <int>,
    "hallucination": <int>
  \},
  "flags": "...",
  "rationale": "...",
  "recommendation": "keep/review/discard"
\}
}
\end{promptbox}
\vspace{0.5em}

\subsection{Diversity Categorization Prompt}
\label{sec:appendix_diversity}

To systematically measure the semantic diversity of candidate datasets, we employ an LLM-based classifier to map each data sample to a precise primary category and subcategory. This process ensures our optimization algorithm (Section 2.3) correctly balances the distribution across various multimodal capabilities.

\vspace{0.5em}
\begin{promptbox}[Diversity Classification Prompt]
{\small \ttfamily
\noindent You are an expert classifier specialized in categorizing multimodal (image-text) or text-only data into defined types, aligned with the data structures of state-of-the-art (SOTA) models. Your goal is to accurately map each sample to a primary category and subcategory based on content and image dependency.

\noindent \textbf{Note on Image Dependency:} \\
- \textbf{Multimodal Categories} (Caption, OCR, General VQA, Interleave, STEM, GUI, Grounding): Require an image or <image> placeholder.\\
- \textbf{Text-Only Category}: No image dependency—text contains no <image> placeholder and is fully self-contained.

\noindent \textbf{1. Primary Categories \& Subcategories}
\begin{itemize}[leftmargin=1em]
    \vspace{-0.5em}\item \textbf{Caption:} textual descriptions of single or tightly arranged multiple images (basic\_caption, fine\_grained\_caption, scene\_caption, comparative\_caption, multilingual\_caption).
    \vspace{-0.5em}\item \textbf{OCR:} Extracting and using text from images to solve tasks (text\_detection, text\_recognition, text\_understanding, scene\_text, document\_ocr).
    \vspace{-0.5em}\item \textbf{Grounding:} Linking text to specific visual elements like bbox or video segments (location\_grounding, phrase\_grounding, referring\_expression, counting\_grounding, multimodal\_alignment).
    \vspace{-0.5em}\item \textbf{STEM:} Integrating advanced math/science principles with specialized visuals (scientific\_image\_qa, complex\_math\_problem\_solving, engineering\_schematic).
    \vspace{-0.5em}\item \textbf{Interleave:} Alternating Q\&A interaction between multiple images and text (image\_text\_alignment, image\_guided\_text, text\_guided\_image, multimodal\_dialogue).
    \vspace{-0.5em}\item \textbf{General VQA:} General image/video-based QA not involving Grounding or OCR (factual\_vqa, reasoning\_vqa, open\_domain\_vqa, multi\_turn\_vqa, commonsense\_vqa).
    \vspace{-0.5em}\item \textbf{GUI:} Agent-centric multimodal instructions in interactive interfaces (ui\_navigation, ui\_element\_interaction, ui\_task\_planning, ui\_command\_execution).
    \vspace{-0.5em}\item \textbf{Text-Only:} Self-contained text (plain\_text, structured\_text, textual\_qa, textual\_summarization, textual\_reasoning).
\end{itemize}

\noindent \textbf{Other Rules:} \\
- Takes Priority: text\_only > OCR = interleave = caption > grounding = gui > stem > general\_vqa.\\
- The core distinction between OCR and STEM lies in reasoning depth: direct correspondence belongs to OCR; complex reasoning belongs to STEM.

\noindent \textbf{2. Output Format} \\
Respond strictly in JSON with:
\{
  "tags": \{
    "primary\_category": "<caption/.../text\_only>",
    "subcategory": "<corresponding subcategory>"
  \},
  "flags": "<comma-separated data features>",
  "rationale": "<detailed explanation linking data to category>"
\}
}
\end{promptbox}
\vspace{0.5em}

\subsection{Difficulty Assessment Formulation}
The Difficulty score $D$ identifies samples requiring advanced capabilities. It aggregates six dimensions, with Cross-Modal Synthesis and Prior Knowledge Demand weighted most heavily, as they are key drivers for VLM scaling. The formula is:
\begin{equation}
% \resizebox{\linewidth}{!}{$
    D = \frac{0.5 \cdot S_{txt} + 0.5 \cdot S_{img} + 2 \cdot S_{cross} + 1.5 \cdot S_{prior} + 1 \cdot S_{cue} + 0.3 \cdot S_{amb}}{5.8}
% $}
\end{equation}
where the variables correspond to Text Complexity, Image Complexity, Cross-Modal Synthesis, Prior Knowledge Demand, Visual Cue Sensitivity, and Task Ambiguity.

\vspace{0.5em}
\begin{promptbox}[Difficulty Evaluation Prompt]
{\small \ttfamily
\noindent You are an expert evaluator specialized in screening high-difficulty, valuable multimodal (image-answer) training data for large language models. Your goal is to identify samples requiring advanced capabilities (complex reasoning, cross-modal integration, expert knowledge, precise visual perception) while excluding low-value noise.

\noindent \textbf{1. Core Evaluation Dimensions} \\
Rate 6 key difficulty dimensions on a 1-5 scale. Use 2 or 4 for intermediate difficulty:

\noindent \textbf{a. Image Complexity:} Measures the difficulty of understanding the image on its own.\\
(1: Simple object; 3: Multiple interacting objects; 5: Abstract/hyper-detailed content)

\noindent \textbf{b. Text Complexity:} Measures the difficulty of understanding the answer text.\\
(1: Short phrases; 3: Domain-general sentences; 5: Complex syntax, jargon)

\noindent \textbf{c. Cross-Modal Synthesis:} Measures the difficulty of connecting image and text.\\
(1: No meaningful link; 3: Moderate link, 1-2 steps inference; 5: Deep, abstract link, multi-step)

\noindent \textbf{d. Prior Knowledge Demand:} Measures domain/cultural knowledge required.\\
(1: No prior knowledge; 3: Basic domain knowledge; 5: Cross-domain expert knowledge)

\noindent \textbf{e. Visual Cue Sensitivity:} Measures the difficulty of perceiving task-relevant critical visual cues.\\
(1: Cues are large, high-contrast; 3: Small, low-contrast; 5: Nearly imperceptible, requires magnification)

\noindent \textbf{f. Task Ambiguity:} Measures the difficulty of resolving uncertainty.\\
(1: No ambiguity; 3: Requires close inspection; 5: Requires multi-step expert reasoning)

\noindent \textbf{2. Output Format} \\
Respond strictly in JSON with:
\{
  "dimension\_scores": \{
    "image\_complexity": <int>,
    "text\_complexity": <int>,
    "cross\_modal\_synthesis": <int>,
    "prior\_knowledge\_demand": <int>,
    "visual\_cue\_sensitivity": <int>,
    "task\_ambiguity": <int>
  \},
  "flags": "...",
  "rationale": "..."
\}
}
\end{promptbox}

\section{Detailed Hardware and Hyperparameters}
\label{sec:appendix_training_details}

For Stage 2, we use a global batch size of 128 and a peak learning rate of $1e^{-4}$ with cosine decay. 
We enable sequence packing with a maximum length of 8192 to maximize training efficiency. 
All models are trained using the AdamW~\cite{loshchilov2017decoupled} optimizer.
% All experiments were conducted on a high-performance computing cluster equipped with NVIDIA A100 (80GB) GPUs. 
To handle the memory overhead of the context length, we utilized DeepSpeed ZeRO-2 optimization~\cite{aminabadi2022deepspeed}. 
Our efficient data pipeline and training stack allowed for rapid iteration: training the 10B token recipe in Stage 2 required approximately 480 GPU hours, all of our training experiments totally cost about 3500 GPU hours. 
This efficiency is central to our methodology, proving that systematic data construction enables high-performance VLM training with manageable computational budgets (accessible even within reasonable industrial constraints).

\section{Intra-class Sampling Visualization Analysis}
\label{sec:appendix_vis_intra_sampling}

\begin{figure*}[ht]
    \centering
    \includegraphics[width=\textwidth]{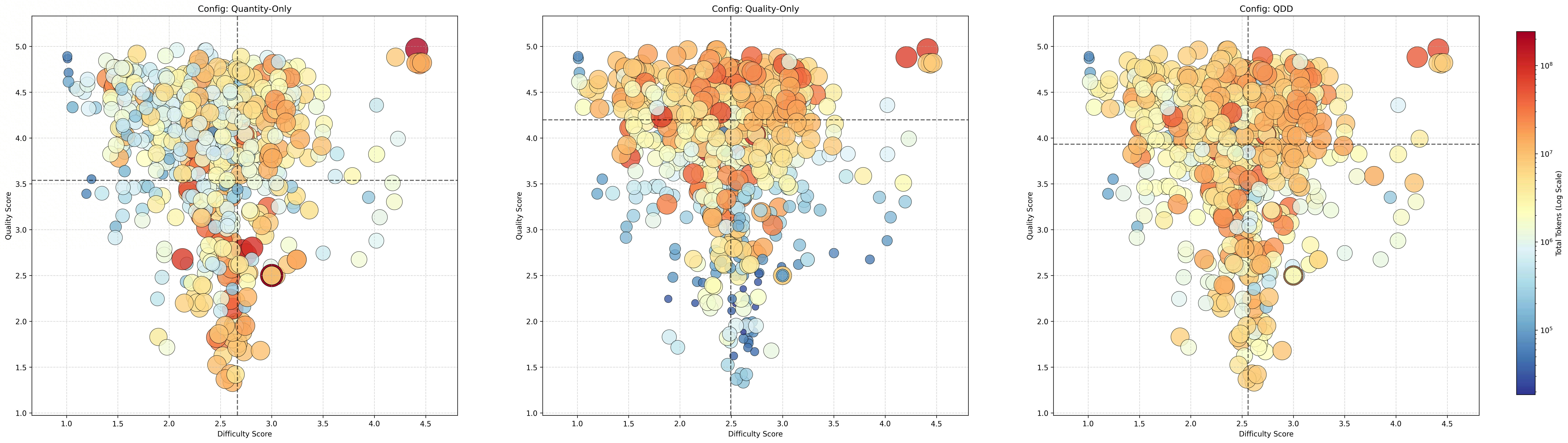} 
    \caption{\textbf{Visualization of Intra-class Sampling Distributions.} The X-axis represents the Difficulty score, and the Y-axis represents the Quality score. The size of each point corresponds to the allocated token volume (log-scaled). From left to right: Quantity-Proportional, Quality-Only, and our convex intra-class allocation.}
    \label{fig:qdd_scatter}
\end{figure*}

Figure~\ref{fig:qdd_scatter} visualizes token allocations across the Quality and Difficulty dimensions for each strategy. Quantity-proportional sampling assigns substantial mass to lower-quality, lower-difficulty regions, whereas quality-only sampling concentrates toward higher Quality scores without explicitly accounting for Difficulty. Our convex allocation shifts more token mass toward datasets that score highly on both dimensions while retaining broader coverage, illustrating the behavior encouraged by its objective.

\begin{table*}[t]
    \centering
    \renewcommand\arraystretch{1.2}
    \setlength{\tabcolsep}{2.5pt}
    \footnotesize
    \resizebox{\textwidth}{!}{
    \begin{tabular}{l|YYYYY|YYY|YYY|YYY|YY|c}
        \toprule
        \multirow{2}{*}{\textbf{Admission Regime}} & \multicolumn{5}{c|}{\textbf{General}} & \multicolumn{3}{c|}{\textbf{K\&H}} & \multicolumn{3}{c|}{\textbf{OCR}} & \multicolumn{3}{c|}{\textbf{M\&L}} & \multicolumn{2}{c|}{\textbf{Video}} & \multirow{2}{*}{\textbf{Avg}} \\
        & 
        \rotatebox{90}{MMBench} & \rotatebox{90}{MME} & \rotatebox{90}{MMStar} & \rotatebox{90}{R.WorldQA} & \rotatebox{90}{MMMU} & 
        \rotatebox{90}{BLINK} & \rotatebox{90}{HalBench} & \rotatebox{90}{AI2D} & 
        \rotatebox{90}{OCRBench} & \rotatebox{90}{InfoVQA} & \rotatebox{90}{CharXiv} & 
        \rotatebox{90}{LogicVista} & \rotatebox{90}{VisuLogic} & \rotatebox{90}{PuzzleVQA} & 
        \rotatebox{90}{V.MME} & \rotatebox{90}{MVBench} & 
        \\
        \midrule
        Baseline (no admission) & 82.6 & 76.9 & 63.1 & 71.2 & 50.2 & 57.1 & 50.8 & 82.5 & 79.7 & 55.7 & 52.2 & 36.2 & 25.0 & 37.0 & 51.9 & 52.1 & 57.8 \\
        Stack & 82.8 & 73.5 & 63.5 & 69.8 & 49.1 & 57.9 & 51.3 & 83.5 & 80.1 & 57.2 & 54.8 & 36.7 & 22.9 & 37.5 & 52.1 & 53.2 & 57.9 \\
        Displace & 81.5 & 75.5 & 61.2 & 65.5 & 48.9 & 56.8 & 51.9 & 82.3 & 79.8 & 56.2 & 53.2 & 38.0 & 21.9 & 41.0 & 52.1 & 53.1 & 57.4 \\
        \rowcolor{lightred}
        \textbf{Ours} & 82.8 & 74.6 & 62.9 & 72.3 & 51.0 & 57.3 & 50.7 & 82.5 & 80.8 & 56.1 & 51.5 & 37.4 & 25.0 & 37.5 & 52.3 & 52.2 & 57.9 \\
        \bottomrule
    \end{tabular}
    }
    \caption{\label{tab:attributable_full}
    \textbf{Full per-benchmark attribution results.} Absolute scores for admitting one OCR dataset under each regime, from which the per-domain $\Delta$ in Table~\ref{tab:attributable} are derived; \emph{Ours} in \colorbox{lightred}{light red}. All runs use the same 4B proxy model and token budget.
    }
\end{table*}

% === 正文中使用 ===
\begin{casebox}[Case 1: Low-Quality OCR Sample]

% --- 上半部分：图片 + QA ---
\begin{minipage}[t]{0.3\columnwidth}
  \centering
  \includegraphics[width=\linewidth]{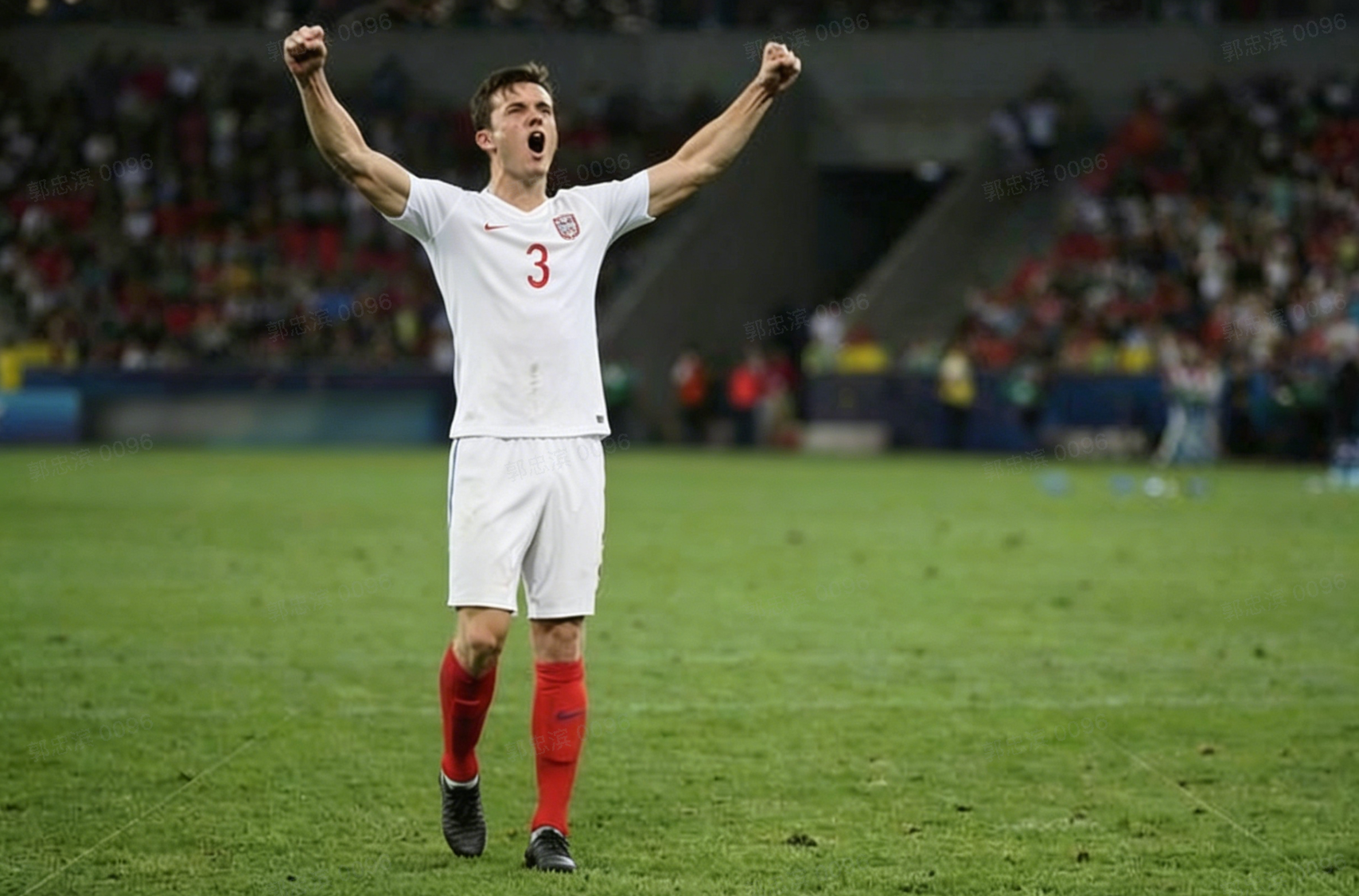}
\end{minipage}%
\hfill
\begin{minipage}[t]{0.66\columnwidth}
  \small
  \vspace{-5em}
  \textbf{Question:} \texttt{<image>} Provide a description of the image content and text. \\[4pt]
  \textbf{Answer:} person celebrates his penalty during soccer league, match \texttt{<ocr>} \texttt{</ocr>}
\end{minipage}

% \vspace{6pt}
\noindent\rule{\linewidth}{0.3pt}
% \vspace{4pt}

% --- 下半部分：Human vs. Model ---
\begin{minipage}[t]{0.42\columnwidth}
  \small
  \textbf{\textcolor{humancolor}{Human Scores}}\\[2pt]
  \begin{tabular}{@{}ll@{}}
    Accuracy & 2 \\
    Correlation & 3 \\
    Grammar & 2 \\
    Hallucination & 1 \\
    \midrule
    \textbf{Overall} & \textbf{1.90} \\
  \end{tabular}
\end{minipage}%
\hfill
\begin{minipage}[t]{0.55\columnwidth}
  \small
  \textbf{\textcolor{modelcolor}{Automated Judge Scores}}\\[2pt]
  \begin{tabular}{@{}ll@{}}
    Accuracy & 2 \\
    Correlation & 3 \\
    Grammar & 3 \\
    Hallucination & 1 \\
    \midrule
    \textbf{Overall} & \textbf{1.96} \\
  \end{tabular}
\end{minipage}

% \vspace{6pt}
\noindent\rule{\linewidth}{0.3pt}
% \vspace{4pt}

% --- Rationale ---
\small
\textbf{\textcolor{modelcolor}{Rationale:}}
\begin{enumerate}[leftmargin=*, itemsep=1pt, parsep=0pt, topsep=2pt, label=\arabic*.]
  \item \textbf{Accuracy}: The response incorrectly claims the player is celebrating a ``penalty,'' which has no basis in the image. While the OCR text is accurate, the penalty assertion is a significant factual error ($\sim$30\%), justifying a score of 2.
  \item \textbf{Grammar}: Misplaced comma (``soccer league , match'') and vague ``person'' instead of ``soccer player'' lead to awkward phrasing---score 3.
  \item \textbf{Correlation}: Relevant elements (celebration, OCR) are present but an irrelevant penalty claim detracts---score 3.
  \item \textbf{Hallucination}: The ``penalty'' claim is a severe hallucination with no visual evidence---score 1.
\end{enumerate}

\end{casebox}

% ----------------------------------------------- %
% === 正文中使用 ===
\begin{casebox}[Case 2: High-Quality General Sample]

% --- 上半部分：图片 + QA ---
\begin{minipage}[t]{0.3\columnwidth}
  \centering
  \includegraphics[width=\linewidth]{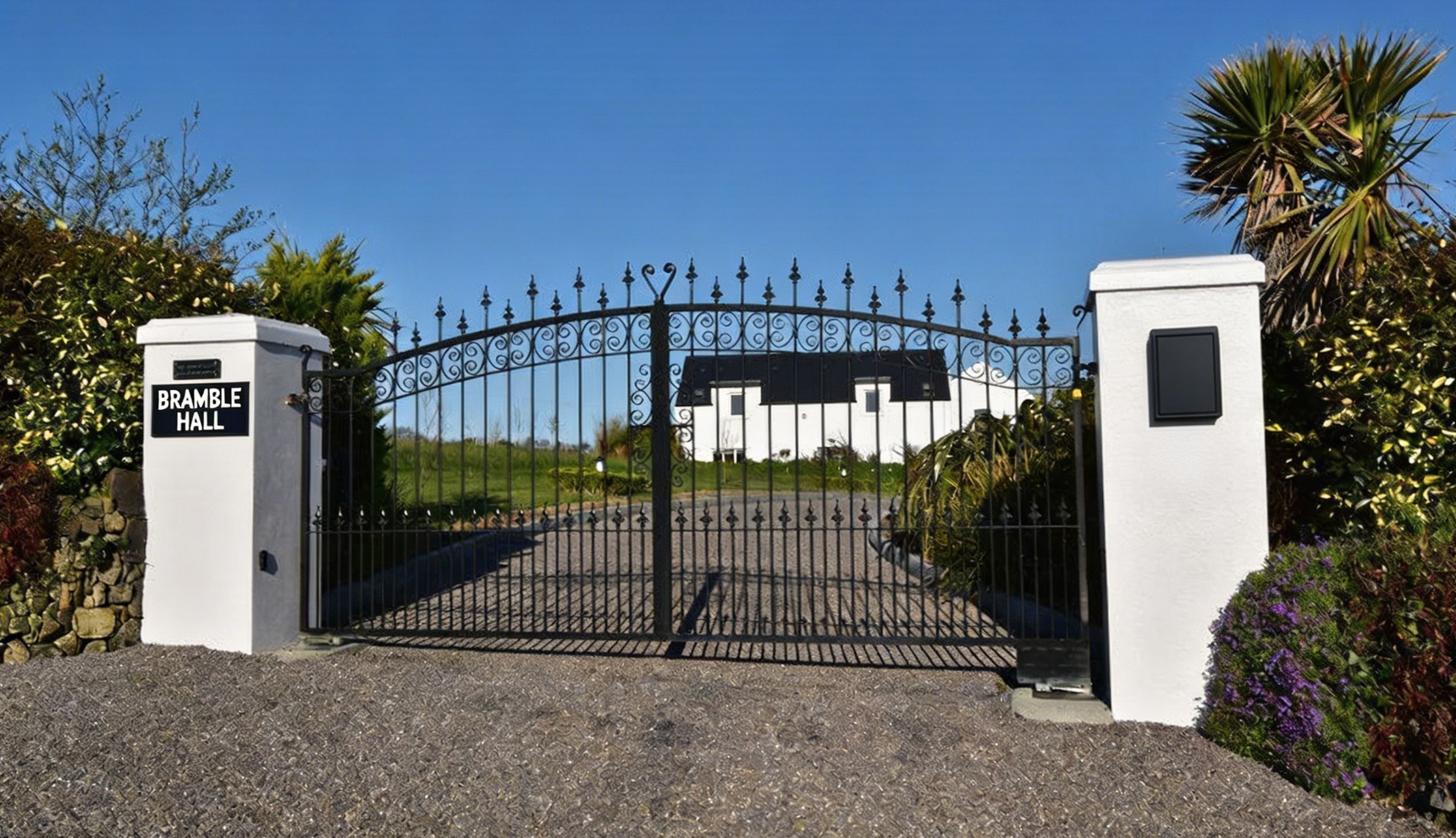}
\end{minipage}%
\hfill
\begin{minipage}[t]{0.66\columnwidth}
  \small
  \vspace{-4em}
  \textbf{Question:} \texttt{<image>} Considering the security measures in place at the entrance, what might be the level of security concern for the property, and what does this imply about the importance or value of ``Bramble Hall''? \\[4pt]
  \textbf{Answer:} The presence of electronic access panels on both sides of the gate suggests a moderate to high level of security concern for the property. This implies that ``Bramble Hall'' is likely to be a property of significance or value, possibly containing assets or individuals that require protection from unauthorized access. 
  % The meticulous attention to security details indicates that the owners prioritize privacy and control over who enters the premises.
\end{minipage}

\vspace{6pt}
\noindent\rule{\linewidth}{0.3pt}
% \vspace{4pt}

% --- 下半部分：Human vs. Model ---
\begin{minipage}[t]{0.42\columnwidth}
  \small
  \textbf{\textcolor{humancolor}{Human Scores}}\\[2pt]
  \begin{tabular}{@{}ll@{}}
    Accuracy & 4 \\
    Correlation & 5 \\
    Grammar & 5 \\
    Hallucination & 4 \\
    \midrule
    \textbf{Overall} & \textbf{4.27} \\
  \end{tabular}
\end{minipage}%
\hfill
\begin{minipage}[t]{0.55\columnwidth}
  \small
  \textbf{\textcolor{modelcolor}{Automated Judge Scores}}\\[2pt]
  \begin{tabular}{@{}ll@{}}
    Accuracy & 4 \\
    Correlation & 5 \\
    Grammar & 5 \\
    Hallucination & 4 \\
    \midrule
    \textbf{Overall} & \textbf{4.27} \\
  \end{tabular}
\end{minipage}

% \vspace{6pt}
\noindent\rule{\linewidth}{0.3pt}
% \vspace{4pt}

% --- Rationale ---
\small
\textbf{\textcolor{modelcolor}{Rationale:}}
% \begin{enumerate}[leftmargin=*, itemsep=1pt, parsep=0pt, topsep=2pt, label=\arabic*.]
%   \item \textbf{Accuracy}: The response incorrectly claims the player is celebrating a ``penalty,'' which has no basis in the image. While the OCR text is accurate, the penalty assertion is a significant factual error ($\sim$30\%), justifying a score of 2.
%   \item \textbf{Grammar}: Misplaced comma (``soccer league , match'') and vague ``person'' instead of ``soccer player'' lead to awkward phrasing---score 3.
%   \item \textbf{Correlation}: Relevant elements (celebration, OCR) are present but an irrelevant penalty claim detracts---score 3.
%   \item \textbf{Hallucination}: The ``penalty'' claim is a severe hallucination with no visual evidence---score 1.
% \end{enumerate}

The assistant correctly identified an electronic access panel on the right gate pillar (consistent with the image) and reasonably inferred moderate-to-high security concern, property significance, and owner prioritization of privacy from the gate and access features. However, the assistant incorrectly claimed electronic access panels are present "on both sides of the gate"—the left pillar only has a ``Bramble Hall'' sign and no access panel, which is a hallucinated detail not supported by the image. This error constitutes ~4\% of the response content. 
% The remaining inferences align with visible evidence (ornate metal gate, secured entrance, private house context).

\end{casebox}

\begin{casebox}[Case 3: Difficulty Assessment -- Grounding Sample]

% --- 上半部分：图片 + QA ---
\begin{minipage}[t]{0.3\columnwidth}
  \centering
  \includegraphics[width=\linewidth]{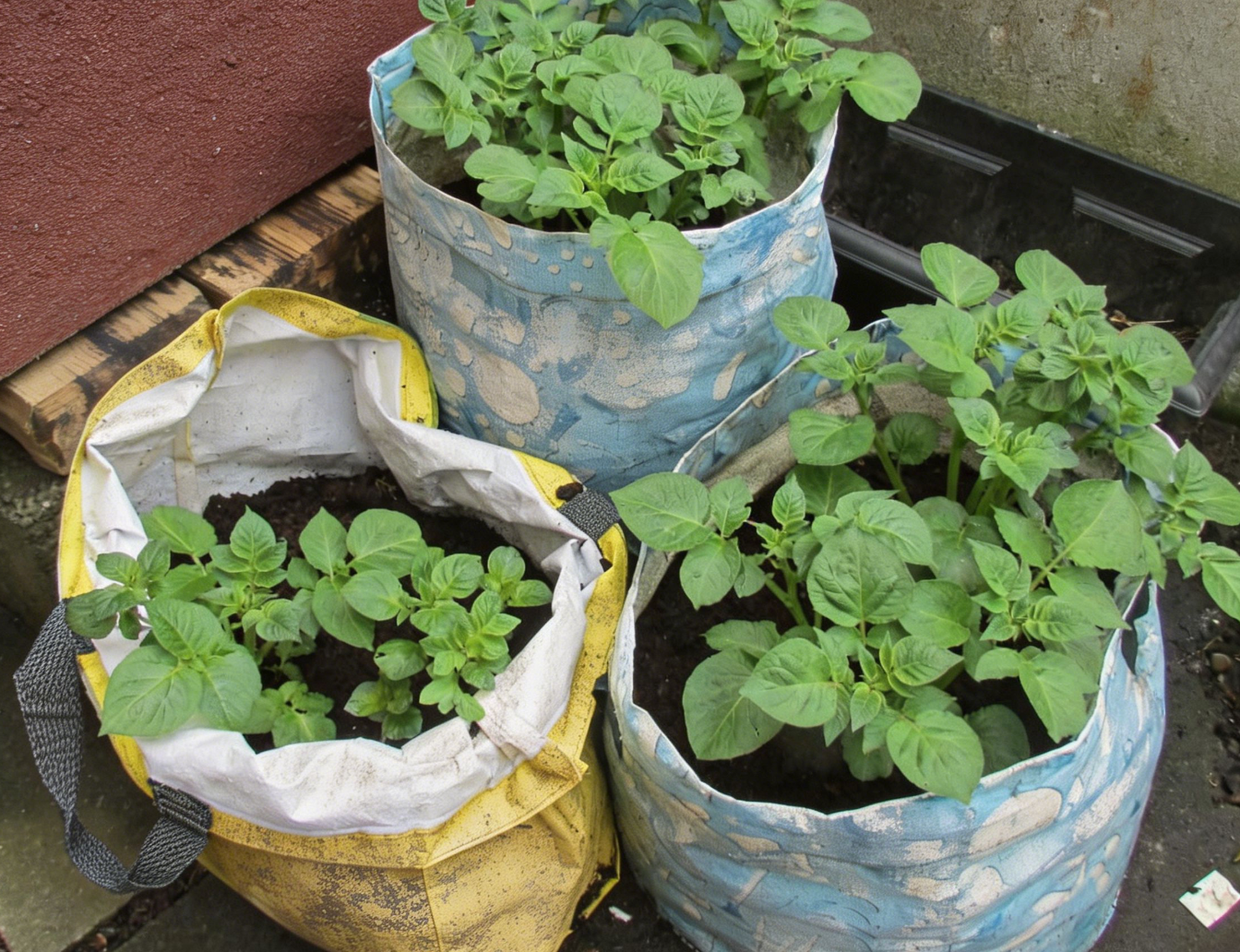}
\end{minipage}%
\hfill
\begin{minipage}[t]{0.66\columnwidth}
  \small
  \vspace{-5.5em}
  \textbf{Question:} \texttt{<image>} Name and localize each object present in the photo. \\[4pt]
  \textbf{Answer:} To grow potatoes in small garden \texttt{<ref-object><bbox>}s, as depicted in the image, place three \texttt{<ref-object><bbox>}s filled with young potato seedlings on a pavement or sidewalk. The \texttt{<ref-object><bbox>}s showcase various designs and colors, including a predominantly yellow and white one, a blue \texttt{<ref-object><bbox>} with a pattern, and a black \texttt{<ref-object><bbox>}. The plants inside the \texttt{<ref-object><bbox>}s are in their early growth \texttt{<ref-object><bbox>}\ldots
\end{minipage}

\vspace{6pt}
\noindent\rule{\linewidth}{0.3pt}
\vspace{4pt}

% % --- 下半部分：Human vs. Model ---
% \begin{minipage}[t]{0.44\columnwidth}
%   \small
%   \textbf{\textcolor{humancolor}{Human Scores}}\\[2pt]
%   \begin{tabular}{@{}lc@{}}
%     Image Complexity & 3 \\
%     Text Complexity & 2 \\
%     Cross-Modal Synthesis & 3 \\
%     Prior Knowledge & 3 \\
%     Visual Cue Sensitivity & 2 \\
%     Task Ambiguity & 5 \\
%     \midrule
%     \textbf{Average} & \textbf{3.02} \\
%   \end{tabular}
% \end{minipage}%
% \hfill
% \begin{minipage}[t]{0.53\columnwidth}
%   \small
%   \textbf{\textcolor{modelcolor}{Automated Judge Scores}}\\[2pt]
%   \begin{tabular}{@{}lc@{}}
%     Image Complexity & 3 \\
%     Text Complexity & 3 \\
%     Cross-Modal Synthesis & 3 \\
%     Prior Knowledge & 3 \\
%     Visual Cue Sensitivity & 1 \\
%     Task Ambiguity & 5 \\
%     \midrule
%     \textbf{Average} & \textbf{2.84} \\
%   \end{tabular}
% \end{minipage}
\centering
\small
\begin{tabular}{@{}l c c@{}}
  \toprule
  \textbf{Dimension} & \textbf{\textcolor{humancolor}{Human}} & \textbf{\textcolor{modelcolor}{Auto Judge}} \\
  \midrule
  Image Complexity       & 3 & 3 \\
  Text Complexity        & 2 & 3 \\
  Cross-Modal Synthesis  & 3 & 3 \\
  Prior Knowledge        & 3 & 3 \\
  Visual Cue Sensitivity & 2 & 1 \\
  Task Ambiguity         & 5 & 5 \\
  \midrule
  \textbf{Average}       & \textbf{3.02} & \textbf{2.84} \\
  \bottomrule
\end{tabular}

\vspace{6pt}
\noindent\rule{\linewidth}{0.3pt}
\vspace{4pt}

% --- Rationale ---
\small
\textbf{\textcolor{modelcolor}{Rationale:}}
\begin{itemize}[leftmargin=*, itemsep=1pt, parsep=0pt, topsep=2pt]
  \item \textbf{Image Complexity}=3: Multiple interacting objects (three grow bags, potato plants, background elements like a red wall, wood, and concrete)---moderate detail, no abstract content.
  \item \textbf{Text Complexity}=3: Moderate-length sentences using domain-general gardening terms like ``young potato seedlings'' and ``early growth stage.''
  \item \textbf{Cross-Modal Synthesis}=3: Links image elements (colored grow bags, green plants) to gardening context, requiring 1--2 inference steps to recognize bags as planters and plants as potato seedlings.
  \item \textbf{Prior Knowledge}=3: Basic gardening knowledge needed to identify grow bags, potato seedlings, and understand early growth stages.
  \item \textbf{Visual Cue Sensitivity}=1: Task-relevant cues (grow bags and plants) are large, high-contrast, and immediately noticeable.
  \item \textbf{Task Ambiguity}=5: No uncertainty in naming/locating core objects; the task is straightforward.
\end{itemize}

\end{casebox}

\end{document}